\documentclass{article}

\usepackage[preprint, nonatbib]{neurips_2022}

\usepackage{graphicx}
\usepackage[utf8]{inputenc} 
\usepackage[T1]{fontenc}    
\usepackage{hyperref}       
\usepackage{url}            
\usepackage{booktabs}       
\usepackage{amsfonts}       
\usepackage{amssymb}
\usepackage{nicefrac}       
\usepackage{microtype}      
\usepackage[table]{xcolor} 
\usepackage{subcaption}
\usepackage{multirow} 
\usepackage{soul}
\usepackage{wrapfig}
\usepackage{algorithm}
\usepackage{algpseudocode}

\graphicspath{ {images/} {images-suppl/} }

\title{Smart Multi-tenant Federated Learning}

\author{Weiming Zhuang$^{1,3}$, \, Yonggang Wen$^{2}$, \, Shuai Zhang$^{3}$ \\
$^{1}$S-Lab, NTU, Singapore\, $^{2}$NTU, Singapore\, $^{3}$SenseTime Research \\
\texttt{weiming001@e.ntu.edu.sg,ygwen@ntu.edu.sg,zhangshuai@sensetime.com}
}

\begin{document}

\maketitle

\begin{abstract}
Federated learning (FL) is an emerging distributed machine learning method that empowers in-situ model training on decentralized edge devices. However, multiple simultaneous training activities could overload resource-constrained devices. In this work, we propose a smart multi-tenant FL system, MuFL, to effectively coordinate and execute simultaneous training activities. We first formalize the problem of multi-tenant FL, define multi-tenant FL scenarios, and introduce a vanilla multi-tenant FL system that trains activities sequentially to form baselines. Then, we propose two approaches to optimize multi-tenant FL: 1) \emph{activity consolidation} merges training activities into one activity with a multi-task architecture; 2) after training it for rounds, \emph{activity splitting} divides it into groups by employing affinities among activities such that activities within a group have better synergy. Extensive experiments demonstrate that MuFL outperforms other methods while consuming 40\% less energy. We hope this work will inspire the community to further study and optimize multi-tenant FL.

\end{abstract}

\section{Introduction}
Federated learning (FL) \cite{fedavg} has attracted considerable attention as it enables privacy-preserving distributed model training among decentralized devices. It is empowering growing numbers of applications in both academia and industry, such as Google Keyboard \cite{hard2018gboard}, medical imaging analysis \cite{li2019brain-tumor1,sheller2018brain-tumor2}, and autonoumous vehicles \cite{zhang2021auto,nguyen2021auto1,posner2021vehicular}. Among them, some applications contain multiple training activities for different tasks. For example, Google Keyboard includes query suggestion \cite{yang2018gboard}, emoji prediction \cite{ramaswamy2019emoji}, and next-world prediction \cite{hard2018gboard}; autonomous vehicles relates to multiple computer vision (CV) tasks, including object detection, tracking, and semantic segmentation \cite{janai2020cvforauto}. 

However, multiple simultaneous training activities could overload edge devices \cite{bonawitz2019flsys}. Edge devices have tight resource constraints, whereas training deep neural networks for the aforementioned applications is resource-intensive. As a result, the majority of edge devices can only support one training activity at a time \cite{liu2019performance}; multiple simultaneous federated learning activities on the same device could overwhelm its memory, computation, and power capacities.
Thus, it is important to navigate solutions to well coordinate these training activities.

A plethora of research on FL considers only one training activity in an application. Many studies are devoted to addressing challenges including statistical heterogeneity \cite{fedprox, wang2020fedma,wang2020fednova, zhuang2020fedreid, deng2021adaptive, zhang2021parameterized}, system heterogeneity \cite{chai2020tifl, yang2021characterizing}, communication efficiency \cite{fedavg, jakub2016communication, karimireddy2020scaffold, zhu2021delayed}, and privacy issues \cite{bagdasaryan2020backdoor, huang2021evaluating}. A common limitation is that they only focus on one training activity, but applications like Google Keyboard require multiple training activities for different targets \cite{yang2018gboard,hard2018gboard,ramaswamy2019emoji}. Multi-tenancy of an FL system is designed in \cite{bonawitz2019flsys} to prevent simultaneous training activities from overloading devices. However, it mainly considers differences among training activities, neglecting potential synergies.

In this work, we propose a smart multi-tenant federated learning system, MuFL, to efficiently coordinate and execute simultaneous training activities under resource constraints by considering both synergies and differences among training activities. We first formalize the problem of multi-tenant FL and define four multi-tenant FL scenarios based on two variances in Section \ref{sec:problem-setup}: 1) whether all training activities are the same type of application, e.g., CV applications; 2) whether all clients support all training activities. Then, we define a vanilla multi-tenant FL system that supports all scenarios by training activities sequentially. Built on it, we further optimize the scenario, where all training activities are the same type and all clients support all activities, by considering both synergies and differences among activities in Section \ref{sec:smart-multitenant-fl}. Specifically, we propose \emph{activity consolidation} to merge training activities into one activity with a multi-task architecture that shares common layers and has specialized layers for each activity. We then introduce \emph{activity splitting} to divide the activity into multiple activities based on their synergies and differences measured by affinities between activities.

We demonstrate that MuFL reduces the energy consumption by over 40\% while achieving superior performance to other methods via extensive experiments on three different sets of training activities in Section \ref{sec:experiments}. We believe that this work will inspire the community to further investigate and optimize multi-tenant FL. We summarize our contributions as follows:

\begin{itemize}
  \item We formalize the problem of multi-tenant FL and define four multi-tenant FL scenarios. To the best of our knowledge, we are the first work that investigates multi-tenant FL in-depth.
  \item We propose MuFL, a smart multi-tenant federated learning system to efficiently coordinate and execute simultaneous training activities by proposing activity consolidation and activity splitting to consider both synergies and differences among training activities.
  \item We establish baselines for multi-tenant FL and demonstrate that MuFL elevates performance with significantly less energy consumption via extensive empirical studies. 
\end{itemize}

\section{Related Work}
\label{sec:related-work}

In this section, we first review the concept of multi-tenancy in cloud computing and machine learning. Then, we provide a literature review of multi-task learning and federated learning.



\textbf{Multi-tenancy of Cloud Computing and Machine Learning} \, Multi-tenancy has been an important concept in cloud computing. It refers to the software architecture where a single instance of software serves multiple users \cite{chong2006architecture,fehling2010framework}. Multi-tenant software architecture is one of the foundations of software as a service (SaaS) applications \cite{mietzner2008defining,krebs2012architectural,cai2013saas}. Recently, researchers have adopted this idea to machine learning (especially deep learning) training and inference. Specifically, some studies investigate how to share GPU clusters to multiple users to train deep neural networks (DNN) \cite{jeon2019analysis,zhao2020hived,lao2021atp}, but these methods are for GPU clusters that have enormous computing resources, which are inapplicable to edge devices that have limited resources. Targeting on-device deep learning, some researchers define multi-tenant as processing multiple computer vision (CV) applications for multiple \emph{concurrent} tasks \cite{fang2018nestdnn,jiang2018mainstream}. However, they focus on the multi-tenant on-device \emph{inference} rather than training. 
On the contrary, we focus on multi-tenant federated learning (FL) \emph{training} on devices, where the multi-tenancy refers to multiple concurrent FL training activities. 

\textbf{Multi-task Learning} \, Multi-task learning is a popular machine learning approach to learn models that generalize on multiple tasks \cite{thrun1995learning,zhang2021survey}. A plethora of studies investigate parameter sharing approaches that share common layers of a similar architecture \cite{caruana1997multitask,eigen2015predicting,bilen2016integrated,nekrasov2019real}. Besides, many studies employ new techniques to address the negative transfer problem \cite{kang2011learning,zhao2018modulation} among tasks, including soft parameter sharing \cite{duong2015low,misra2016cross}, neural architecture search \cite{lu2017fully,huang2018gnas,vandenhende2019branched,guo2020learning,sun2020adashare}, and dynamic loss reweighting strategies \cite{kendall2018multi,chen2018gradnorm,yu2020gradient}. Instead of training all tasks together, task grouping trains only similar tasks together. The early works of task grouping \cite{kang2011learning,kumar2012learning} are not adaptable to DNN. Recently, several studies analyze the task similarity \cite{standley2020which} and task affinities \cite{fifty2021tag} for task grouping. In this work, we adopt the idea of task grouping to consolidate and split training activities. The state-of-the-art task grouping methods \cite{standley2020which,fifty2021tag}, however, are unsuitable for our scenario because they focus on the inference efficiency, bypassing the intensive computation on training. Thus, we propose activity consolidation and activity splitting to group training activities based on their synergies and differences.


\textbf{Federated Learning} \, Federated learning emerges as a promising privacy-preserving distributed machine learning technique that uses a central server to coordinate multiple decentralized clients to train models \cite{fedavg,kairouz2021advances}. The majority of studies aim to address the challenges of FL, including statistical heterogeneity \cite{fedprox, wang2020fedma, wang2020fednova, zhuang2021fedureid, zhang2021parameterized, zhuang2021fedu, zhuang2022fedema, zhuang2022fedreid}, system heterogeneity \cite{chai2020tifl, yang2021characterizing}, communication efficiency \cite{fedavg, jakub2016communication, karimireddy2020scaffold, zhu2021delayed}, and privacy concerns \cite{bagdasaryan2020backdoor, huang2021evaluating}. Among them, federated multi-task learning \cite{smith2017fedmultil,marfoq2021federated} is an emerging method to learn personalized models to tackle the statistical heterogeneity. However, these works mainly focus on training one activity in a client of an application. Multi-tenant FL that handles multiple concurrent training activities is rarely discussed \cite{bonawitz2019flsys}. In this work, we formalize the problem of multi-tenant FL, define four multi-tenant FL scenarios, and optimize one of the scenarios by considering both synergies and differences among training activities.


\section{Problem Setup}
\label{sec:problem-setup}

This section provides preliminaries of federated learning (FL), presents the problem definition of multi-tenant FL, and classifies four multi-tenant FL scenarios. Besides, we introduce a vanilla multi-tenant FL system supports for all scenarios. 

\subsection{Preliminaries and Problem Definition}

In the federated learning setting, the majority of studies consider optimizing the following problem:

\begin{equation}
  \min_{\omega \in \mathbb{R}^d} f(\omega) := \sum_{k=1}^K p_k f_k(\omega) := \sum_{k=1}^K p_k \mathbb{E}_{\xi_k \sim \mathcal{D}_k}[f_k(\omega;\xi_k)],
  \label{eq:fl}
\end{equation}

where $\omega$ is the optimization variable, $K$ is the number of selected clients to execute training, $f_k(\omega)$ is the loss function of client $k$, $p_k$ is the weight of client $k$ in model aggregation, and $\xi_k$ is the training data sampled from data distribution $\mathcal{D}_k$ of client $k$. \verb+FedAvg+ \cite{fedavg} is a popular federated learning algorithm, which sets $p_k$ to be proportional to the dataset size of client $k$. 

Equation \ref{eq:fl} illustrates the objective of single training activity in FL, but in real-world scenarios, multiple simultaneous training activities could overload edge devices. We further formalize the problem of multi-tenant FL as follows.



In multi-tenant FL, a server coordinates a set of clients $\mathcal{C}$ to execute a set of $n$ FL training activities $\mathcal{A} = \{\alpha_1,\alpha_2,\dots,\alpha_n\}$. It obtain a set of parameters of models $\mathcal{W} = \{\omega_1,\omega_2,\dots,\omega_n\}$, where each model $\omega_i$ is for activity $\alpha_i$. By defining $\mathcal{M}(\alpha_i;\omega_i)$ as performance measurement of each training activity $\alpha_i$, multi-tenant FL aims to maximize the performance of all training activities $\sum_{i=1}^n \mathcal{M}(\alpha_i;\omega_i)$, under the constraint that each client $k$ has limited memory budget and computation budget. These budgets constrain the number of concurrent training actvitities $n_k$ on client $k$. Besides, as devices have limited battery life, we would like to minimize the energy consumption and training time to obtain $\mathcal{W}$ from training activities $\mathcal{A}$.



\subsection{Multi-tenant FL Scenarios}

We classify multi-tenant FL into four different scenarios based on variances in two aspects: 1) whether all training activities in $\mathcal{A}$ are the same type of application, e.g., computer vision (CV) applications or natural language processing (NLP) applications; 2) whether all clients in $\mathcal{C}$ support all training activities in $\mathcal{A}$. We depict these four scenarios in Figure \ref{fig:scenarios} in Appendix \ref{apx:scenarios} and describe them below.

\textbf{Scenario 1} \, $\forall \alpha_i \in \mathcal{A}$, $\alpha_i$ is the same type of application; $\forall \alpha_i \in \mathcal{A}$, $\forall c_k \in \mathcal{C}$ supports $\alpha_i$. For example, autonoumous vehicles (clients) support the same sets of CV applications, such as object detection and semantic segmentation. Thus, they support training activities of these applications.

\textbf{Scenario 2} \, $\exists \alpha_i \in \mathcal{A}$, $\alpha_i$ is a different type of application; $\forall \alpha_i \in \mathcal{A}$, $\forall c_k \in \mathcal{C}$ supports $\alpha_i$. For example, Google Keyboard has different types of applications, including recommendation like query suggestion \cite{yang2018gboard} and NLP like next-world prediction \cite{hard2018gboard}. Smartphones (clients) installed Google Keyboard support these applications, thus, supporting all related training activities. 

\textbf{Scenario 3} \, $\forall \alpha_i \in \mathcal{A}$, $\alpha_i$ is the same type of application; $\forall \alpha_i \in \mathcal{A}$, $\exists c_k \in \mathcal{C}$ does not support $\alpha_i$. For example, survelliance cameras (clients) in parking lots could support CV applications, but cameras in different locations may support different applications, e.g., counting open-parking spots, tracking parking duration, or recording fender benders.

\textbf{Scenario 4} \, $\exists \alpha_i \in \mathcal{A}$, $\alpha_i$ is a different type of application; $\forall \alpha_i \in \mathcal{A}$, $\exists c_k \in \mathcal{C}$ does not support $\alpha_i$. For example, browsers (clients) could leverage users' browsing history to support browsing history suggestion \cite{hartmann2019mozilla} and recommendations \cite{minto2021brave}, which are different types of applications. Some users may opt-out the recommendations, resulting in not all browsers need to support all training activities. 

The application determines the multi-tenant FL scenario. We next introduce a vanilla multi-tenant FL that supports all these scenarios as our baseline. 

\subsection{Vanilla Multi-tenant FL}

\begin{figure}[t!]
  \centering
  \begin{subfigure}[t]{0.43\textwidth}
      \includegraphics[width=\textwidth]{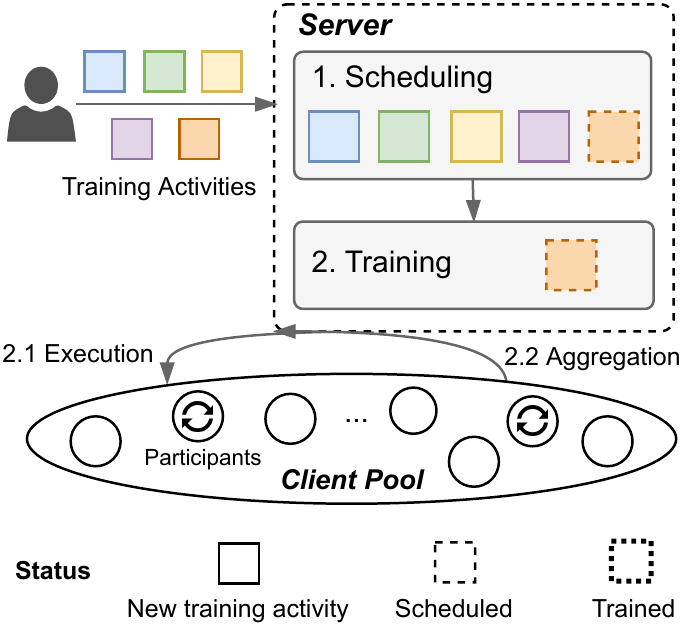}
      \caption{Vanilla Multi-tenant FL}
      \label{fig:vanilla-mtfl}
  \end{subfigure}
  \hfill
  \begin{subfigure}[t]{0.52\textwidth}
     \includegraphics[width=\textwidth]{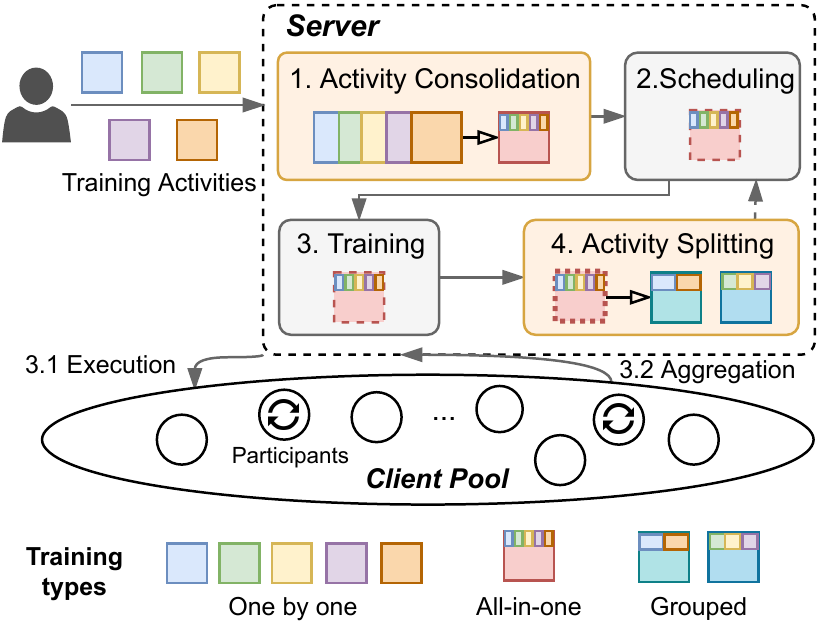}
     \caption{Smart Multi-tenant FL}
     \label{fig:smart-mtfl}
 \end{subfigure}
 \caption{The architectures of proposed multi-tenant federated learning (FL) systems. The vanilla multi-tenant FL system (a) employs a scheduler to queue simultaneous training activities and execute them \emph{one by one}. The smart multi-tenant FL system (b) proposes activity consolidation and activity splitting to consider both synergies and differences among training activities, which elevates performance and reduces resource consumption.}
 \label{fig:mtfl}
\end{figure}

Figure \ref{fig:vanilla-mtfl} presents the architecture of a vanilla multi-tenant FL system. It prevents overloading and congestion of multiple simultaneous training activities by scheduling them to execute \emph{one by one}. Particularly, we design a scheduler to queue training activities in the server. In each round, the server selects $K$ clients from the client pool to participate in training. Depending on the computational resources of the selected clients, the server can execute training for multiple activities concurrently if the clients have enough computation resources. In this study, we assume that each client can execute one training activity at a time ($n_k$ = 1). This is a realistic assumption for the majority of current edge devices. \footnote{Edges devices, e.g., NVIDIA Jetson TX2 and AGX Xavier, have only one GPU; GPU virtualization techniques \cite{dowty2009gpu, hong2017gpu} to enable concurrent training on the same GPU currently are mainly for the cloud stack.} As a result, the vanilla multi-tenant FL system executes training activities sequentially.




The vanilla multi-tenant FL system supports the four multi-tenant FL scenarios discussed above. From the perspective of the type of application, it can handle different application types of training activities as each training activity is executed independently. From the perspective of whether clients support all training activities, each training activity can select clients that support the activity to participate in training. Despite its comprehensiveness, it only considers differences among training activities, neglecting their potential synergies. In contrast, our proposed MuFL considers both synergies and differences among training activities to further optimize the Scenario 1. MuFL can be view as a generalization of the vanilla multi-tenant FL system; it can be specialized to the vanilla multi-tennat FL system by disabling activity consolidation and activity splitting. It is also adaptable to Scenario 3 and 4 by employing strategies to filter out clients that do not support all training activities, but we keep them for future investigation.

\section{Smart Multi-tenant FL}
\label{sec:smart-multitenant-fl}

In this section, we introduce the smart multi-tenant FL system, MuFL. We start by providing an overview of MuFL. Then, we present two important components of MuFL, activity consolidation and activity splitting, to consider both synergies and differences among simultaneous training activities.

Figure \ref{fig:smart-mtfl} depicts the architecture and training processes of MuFL. It contains a server to coordinate training activities and a pool of clients to execute training. MuFL optimizes the Scenario 1 discussed previously with the following steps: 1) The server receives training activities $\mathcal{A} = \{\alpha_1,\alpha_2,\dots,\alpha_n\}$ to train models $\mathcal{W} = \{\omega_1,\omega_2,\dots,\omega_n\}$ and \emph{consolidates} these activities into an all-in-one training activity $\alpha_0$; 2) The server schedules $\alpha_0$ to train; 3) The server select $K$ clients from the client pool to execute $\alpha_0$ iteratively through FL process for $R_0$ rounds; 4) The server \emph{splits} the all-in-one activity $\alpha_0$ into multiple training activity groups $\{\mathcal{A}_1,\mathcal{A}_2,\dots\}$, where each group trains nonoverlapping subset of $\mathcal{W}$. 5) The server iterates step 2 and 3 to train $\mathcal{A}_j$. We summarize MuFL in Algorithm \ref{algo:mufl} in Appendix \ref{apx:algorithm} and introduce the details of activity consolidation and activity splitting next.

\subsection{Activity Consolidation}

Focusing on optimizing the Scenario 1 of multi-tenant FL, we first propose activity consolidation to consolidate multiple training activities into an all-in-one training activity, as illustrated in the first step of Figure \ref{fig:smart-mtfl}. In Scenario 1, all training activities are the same type of application and all clients support all training activities. Since training activities $\mathcal{A} = \{\alpha_1,\alpha_2,\dots,\alpha_n\}$ are of the same type, e.g., CV or NLP, models $\mathcal{W} = \{\omega_1,\omega_2,\dots,\omega_n\}$ could share the same backbone (common layers). Thus, we can consolidate $\mathcal{A}$ into an all-in-one training activity $\alpha_0$ that trains a multi-task model $\nu = \{\theta_s\} \cup \{\theta_{\alpha_i} | \alpha_i \in \mathcal{A} \}$, where $\theta_s$ is the shared model parameters and $\theta_{\alpha_i}$ is the specific parameters for training activity $\alpha_i \in \mathcal{A}$.

\begin{figure}[t!]
  \centering
  \begin{subfigure}[t]{0.49\textwidth}
      \includegraphics[width=\textwidth]{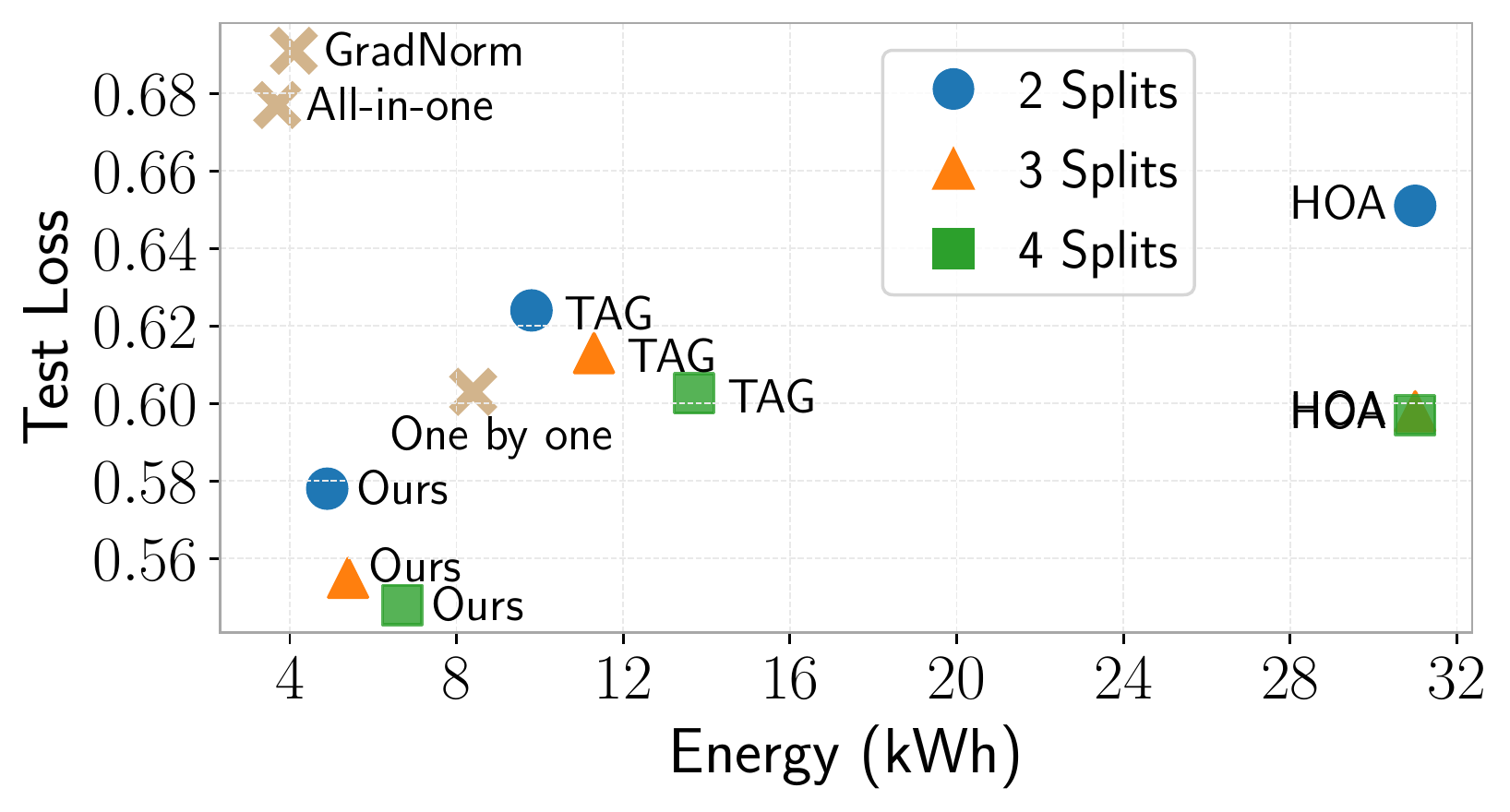}
      \caption{A set of five training activities: \texttt{sdnkt}}
      \label{fig:sdnkt-resource}
  \end{subfigure}
  \hfill
  \begin{subfigure}[t]{0.49\textwidth}
     \includegraphics[width=\textwidth]{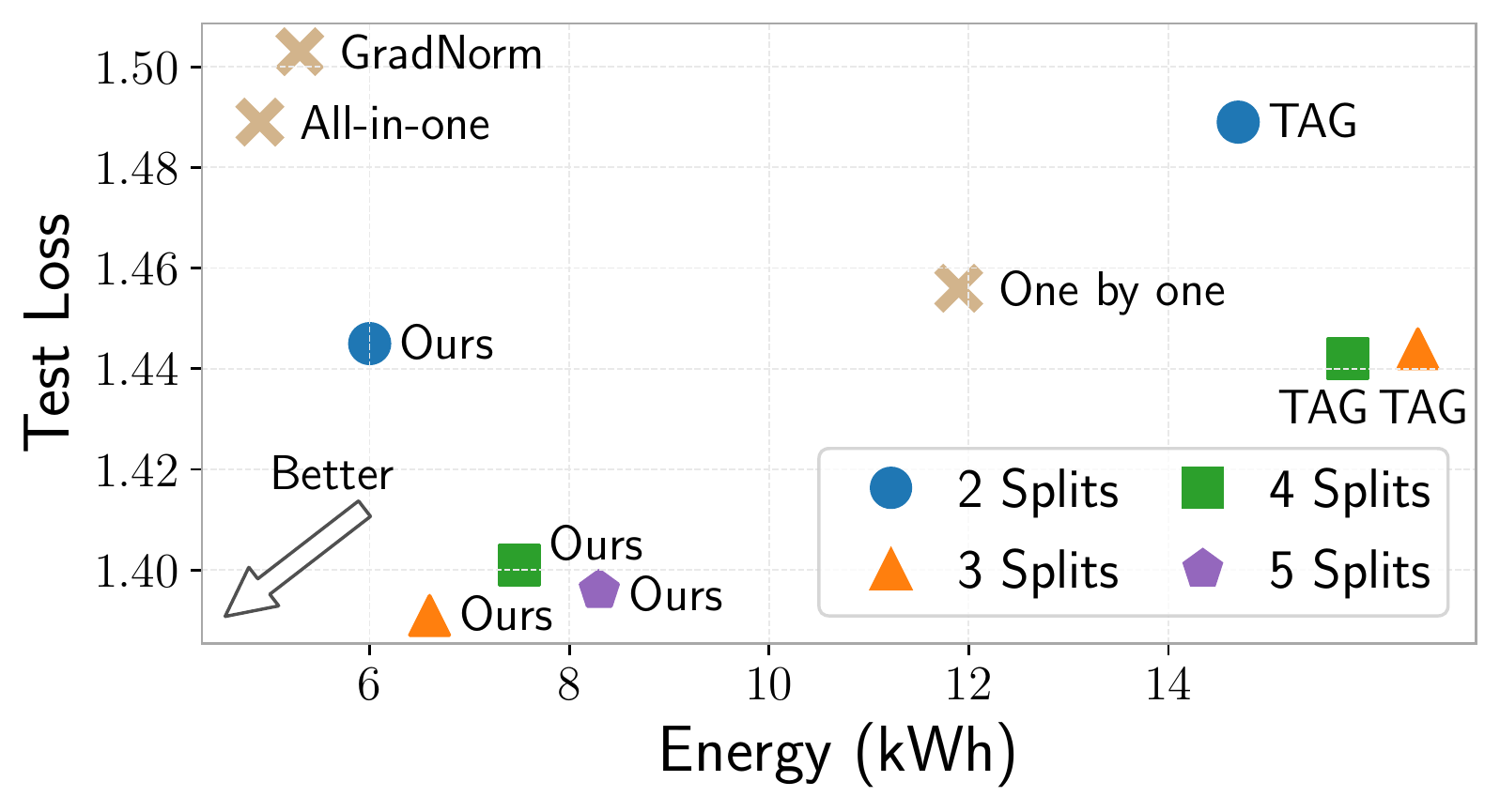}
     \caption{A set of nine trianing activities: \texttt{sdnkterca}}
     \label{fig:sdnkterca-resource}
 \end{subfigure}
 \caption{Comparison of test loss and energy consumption on two training activity sets: (a) \texttt{sdnkt} and (b) \texttt{sdnkterca}, where each character represents an activity. Compared with all-in-one methods, our method achieves much better performance with slight increases in computation. Moreover, our method achieves the best performance while consuming less energy than the other methods.}
 \label{fig:performance-vs-resource}
\end{figure}

Activity consolidation leverages synergies among training activities and effectively reduces the computation cost of multi-tenant FL from multiple trainings into a single training. However, simply employing activity consolidation is another extreme of multi-tenant FL that only considers synergies among activities. As shown in Figure \ref{fig:performance-vs-resource}, all-in-one method is efficient in energy consumption, but it leads to unsatisfactory performance. Consequently, we further propose activity splitting to consider both synergies and differences among training activities.




\subsection{Activity Splitting}

We propose activity splitting to divide the all-in-one activity $\alpha_0$ into multiple groups after it is trained for certain rounds. Essentially, we aim to split $\mathcal{A} = \{\alpha_1,\alpha_2,\dots,\alpha_n\}$ into multiple nonoverlapping groups such that training activities within a group have better synergy. Let $\{\mathcal{A}_1, \mathcal{A}_2,\dots,\mathcal{A}_m\}$ be subsets of $\mathcal{A}$, we aim to find a set $I \subseteq \{1, 2, \dots, m\}$ such that $|I| \leq |\mathcal{A}|$, $\bigcup_{j\in I}\mathcal{A}_j = \mathcal{A}$, and $\bigcap_{j\in I}\mathcal{A}_j = \emptyset$. Each group $\mathcal{A}_j$ trains a model $\nu_j = \{\theta_{s}^{j}\} \cup \{\theta_{\alpha_i} | \alpha_i \in \mathcal{A}_j\}$, which is a multi-task network when $\mathcal{A}_j$ contains more than one training activity, where $\theta_s^j$ is the shared model parameters and $\theta_{\alpha_i}$ is the specific parameters for training activity $\alpha_i \in \mathcal{A}_j$. The core question is how to determine set $I$ to split these activities considering their synergies and differences.





Inspired by TAG \cite{fifty2021tag} that measures task affinites for task grouping, we employ affinities between training activities for activity splitting via three stages: 1) we measure affinities among activities during all-in-one training; 2) we select the best combination of splitted training activities based on affinity scores; 3) we continue training each split with its model initialized with parameters obtained from all-in-one training. Particularly, during training of all-in-one activity $\alpha_0$, we measure the affinity of training activity $\alpha_i$ onto $\alpha_j$ at time step $t$ in each client $k$ with the following equation: 
\begin{equation}
  \mathcal{S}^{k,t}_{\alpha_i \rightarrow \alpha_j} = 1 - \frac{\mathcal{L}_{\alpha_j}(\mathcal{X}^{k,t}, \theta_{s,\alpha_i}^{k,t+1}, \theta_{\alpha_j}^{k,t})}{\mathcal{L}_{\alpha_j}(\mathcal{X}^{k,t}, \theta_s^{k,t}, \theta_{\alpha_j}^{k,t})},
  \label{eq:single-affinity}
\end{equation}
where $\mathcal{L}_{\alpha_j}$ is the loss function of $\alpha_j$, $\mathcal{X}^{k,t}$ is a batch of training data, and $\theta_s^{k,t}$ and $\theta^{k,t+1}_{s,\alpha_i}$ are the shared model parameters \emph{before} and \emph{after} updated by $\alpha_i$, respectively. Positive value of $\mathcal{S}^{k,t}_{\alpha_i \rightarrow \alpha_j}$ means that activity $\alpha_i$ helps reduce the loss of $\alpha_j$. This equation measures the affinity of one time-step of one client. We approximate affinity scores for each round by averaging the values over $T$ time-steps in $E$ local epochs and $K$ selected clients: 
  $\hat{\mathcal{S}}_{\alpha_i \rightarrow \alpha_j} = \frac{1}{K E T} \sum_{k=1}^K \sum_{e=1}^E \sum_{t=1}^T \mathcal{S}^{k,t}_{\alpha_i \rightarrow \alpha_j}$,
where $T$ is total time steps determined by the frequency $f$ of calculating Equation \ref{eq:single-affinity}, e.g., $f = 5$ means measuring the affinity in each client in every five batches.  






These affinity scores measure pair-wise affinities between traininig activities. We next use them to calculate total affinity scores of a grouping with $\sum_{i=1}^n \hat{\mathcal{S}}_{\alpha_i}$, where $\hat{\mathcal{S}}_{\alpha_i}$ is the averaged affinity score onto each training activity. For example, in a grouping of two splits of five training activities $\{ \{\alpha_1, \alpha_2\}, \{\alpha_3, \alpha_4, \alpha_5\} \}$, where $\{\alpha_1, \alpha_2\}$ is one split and $\{\alpha_3, \alpha_4, \alpha_5\}$ is another split. The affinity score onto $\alpha_1$ is $\hat{\mathcal{S}}_{\alpha_1}=\hat{\mathcal{S}}_{\alpha_2 \rightarrow \alpha_1}$ and the affinity score onto $\alpha_3$ is ${\hat{\mathcal{S}}_{\alpha_3}} = (\hat{\mathcal{S}}_{\alpha_4 \rightarrow \alpha_3} + \hat{\mathcal{S}}_{\alpha_5 \rightarrow \alpha_3})/2$. Consequently, we can find the set $I$ with $|I|$ elements for subsets of $\mathcal{A}$ that maximize $\sum_{i=1}^n \hat{\mathcal{S}}_{\alpha_i}$, where $|I|$ defines the number of elements. Although this problem is a NP-hard problem (similar to Set-Cover problem), we can solve it with algorithms like branch-and-bound methods \cite{standley2020which,lawler1966branch}. 

We would like to further highlight the differences between our method and TAG \cite{fifty2021tag}. Firstly, TAG allows overlapping task grouping that could train one task multiple times because it focuses on inference efficiency. In contrast, our focus is fundamentally different; we focus on training efficiency and thus our method considers only nonoverlapping activity splitting. Secondly, TAG is computation-intensive to compute higher numbers of splits, e.g., it fails to produce results of five splits in a week, whereas we only need seconds of computation. Thirdly, TAG sets $\hat{\mathcal{S}}_{\alpha_i \rightarrow \alpha_i} = 1e^{-6}$ in implementation, which rules out the possibility that a group only contains one task as it results in much smaller value than other groupings. Besides, $\hat{\mathcal{S}}_{\alpha_i \rightarrow \alpha_i}$ calculated from Equation \ref{eq:single-affinity} is also not desirable because it is much larger (could be 10x larger) than other affinity scores, resulting in a group that always contains one task. To overcome these issues, we propose a new method to calculate this value:
\begin{equation}
  \hat{\mathcal{S}}_{\alpha_i \rightarrow \alpha_i} = \sum_{j \in \mathcal{N}\backslash\{i\}} \frac{(\hat{\mathcal{S}}_{\alpha_i \rightarrow \alpha_j} + \hat{\mathcal{S}}_{\alpha_j \rightarrow \alpha_i})}{2n-2},
  \label{eq:self-affinity}
\end{equation}
where $\mathcal{N}=\{1,2,\dots,n\}$. The intuition is that it measures the normalized affinity of activity $\alpha_i$ to other activities and other activities to $\alpha_i$. Fourthly, we focus on the multi-tenant FL scenario, thus, we propose to aggregate the affinity scores over $K$ selected clients. Fifthly, TAG trains each set $\mathcal{A}_j$ from scratch, whereas we train each set $\mathcal{A}_j$ by initializing models with the parameters obtained from $R_0$ rounds of all-in-one training.

\section{Experiments}
\label{sec:experiments}

We evaluate the performance and resource usage of MuFL and design our experiments to answer the following questions: 1) How effective is our activity splitting approach? 2) When to split the training activities? 3) Is it beneficial to iteratively split the training activities? 4) What is the impact of local epoch and scaling up the number of selected clients in each training round?

\paragraph{Experiment Setup} We construct the \emph{Scenario 1} of multi-tenant FL scenarios using Taskonomy dataset \cite{zamir2018taskonomy}, which is a large computer vision dataset of indoor scenes of buildings. We run experiments with $N = 32$ clients, where each client contains a dataset of a building to simulate the statistical heterogeneity in FL. Three sets of training activities are used to evaluate the robustness of MuFL: \texttt{sdnkt}, \texttt{erckt}, and \texttt{sdnkterca}; each character represents an activity, e.g., \texttt{s} and \texttt{d} represents semantic segmentation and depth estimation, respectively. We measure the statistical performance of an activity set using the sum of test losses of individual activities. By default, we use $K = 4$ selected clients and $E = 1$ local epoch. More experimental details are provided in Appendix \ref{apx:experimental-setup}.

\subsection{Performance Evaluation}

We compare the performance, in terms of test loss and energy consumption, among the following methods: 1) one by one training of activities (i.e., the vanilla multi-tenant FL); 2) all-in-one training of activities (i.e., using only activity consolidation); 3) all-in-one training with gradient normalization applied to tune the gradient magnitudes among activities (GradNorm \cite{chen2018gradnorm}); 4) estimating higher-order of activity groupings from pair-wise activities performance (HOA \cite{standley2020which}); 5) grouping training activities with only task affinity grouping method (TAG \cite{fifty2021tag}); 6) MuFL with both activity consolidation and activity splitting. Carbontracker \cite{anthony2020carbontracker} is used to measure energy consumption and carbon footprint (provided in Appendix \ref{apx:evaluation}). We report results of multiple splits in activity splitting. 


Figure \ref{fig:performance-vs-resource} compares performance of the above methods on activity sets \texttt{sdnkt} and \texttt{sdnkterca}. The methods that achieve lower test loss and lower energy consumption are better. At the one extreme, all-in-one methods (including GradNorm) consume the least energy, but their test losses are the highest. At the other extreme, HOA \cite{standley2020which} achieves comparable test losses on three or four splits of \texttt{sdnkt}, but it demands high energy consumption ($\sim4-6\times$ of ours) to compute pair-wise activities for higher-order estimation. Although training activities one by one and TAG \cite{fifty2021tag} present a good balance between test loss and energy consumption, MuFL is superior in both aspects; it achieves the best test loss with $\sim$40\% and $\sim$50\% less energy consumption on activity set \texttt{sdnkt} and \texttt{sdnkterca}, respectively. Additionally, more splits of activity in the activity splitting lead to higher energy consumption, but it could help further reduce test losses. We do not report HOA \cite{standley2020which} for activity set \texttt{sdnkterca} due to computation constraints as it requires computing at least 36 pair-wise training activities ($\sim$720 GPU hours); its energy consumption is estimated to be $\sim 12\times$ of ours. We provide test losses of each activity, the results of splits, and results of activity set \texttt{erckt} in Appendix \ref{apx:evaluation}.




\subsection{How Effective is Our Activity Splitting Approach?}

\begingroup
\setlength{\tabcolsep}{0.48em}
\begin{table}[t]\centering
  \caption{Performance (test loss) comparison of our method with the optimal and worst splits. Our method achieves the best performance, indicating the effectiveness of our activity splitting method.}
  \label{tab:comparison-optimal-worst}
  \begin{tabular}{ccccccccc}\toprule
    Activity &\multirow{2}{*}{Splits} &\multirow{2}{*}{Ours} &\multicolumn{2}{c}{Train from Sctrach} & &\multicolumn{2}{c}{Train from Initialization} \\\cmidrule{4-5}\cmidrule{7-8}
    Set & & &Optimal &Worst & &Optimal &Worst \\\midrule
    \multirow{2}{*}{\texttt{sdnkt}} &2 &\textbf{0.578 {\footnotesize ± 0.015}} &0.622 {\footnotesize ± 0.007} &0.685 {\footnotesize ± 0.010} & &0.595 {\footnotesize ± 0.008} &0.595 {\footnotesize ± 0.004} \\
    &3 &\textbf{0.555 {\footnotesize ± 0.008}} &0.585 {\footnotesize ± 0.026} &0.674 {\footnotesize ± 0.022} & &0.560 {\footnotesize ± 0.006} &0.578 {\footnotesize ± 0.006} \\\midrule
    \multirow{2}{*}{\texttt{erckt}} &2 &\textbf{1.039 {\footnotesize ± 0.024}} &1.070 {\footnotesize ± 0.013} &1.312 {\footnotesize ± 0.065} & &1.048 {\footnotesize ± 0.024} &1.068 {\footnotesize ± 0.037} \\
    &3 &\textbf{1.015 {\footnotesize ± 0.018}} &1.058 {\footnotesize ± 0.029} &1.243 {\footnotesize ± 0.099} & &1.020 {\footnotesize ± 0.012} &1.052 {\footnotesize ± 0.026} \\
    \bottomrule
  \end{tabular}
\end{table}
\endgroup

\begin{figure}[t]
  \centering
  \begin{subfigure}[t]{0.31\textwidth}
    \includegraphics[width=\textwidth]{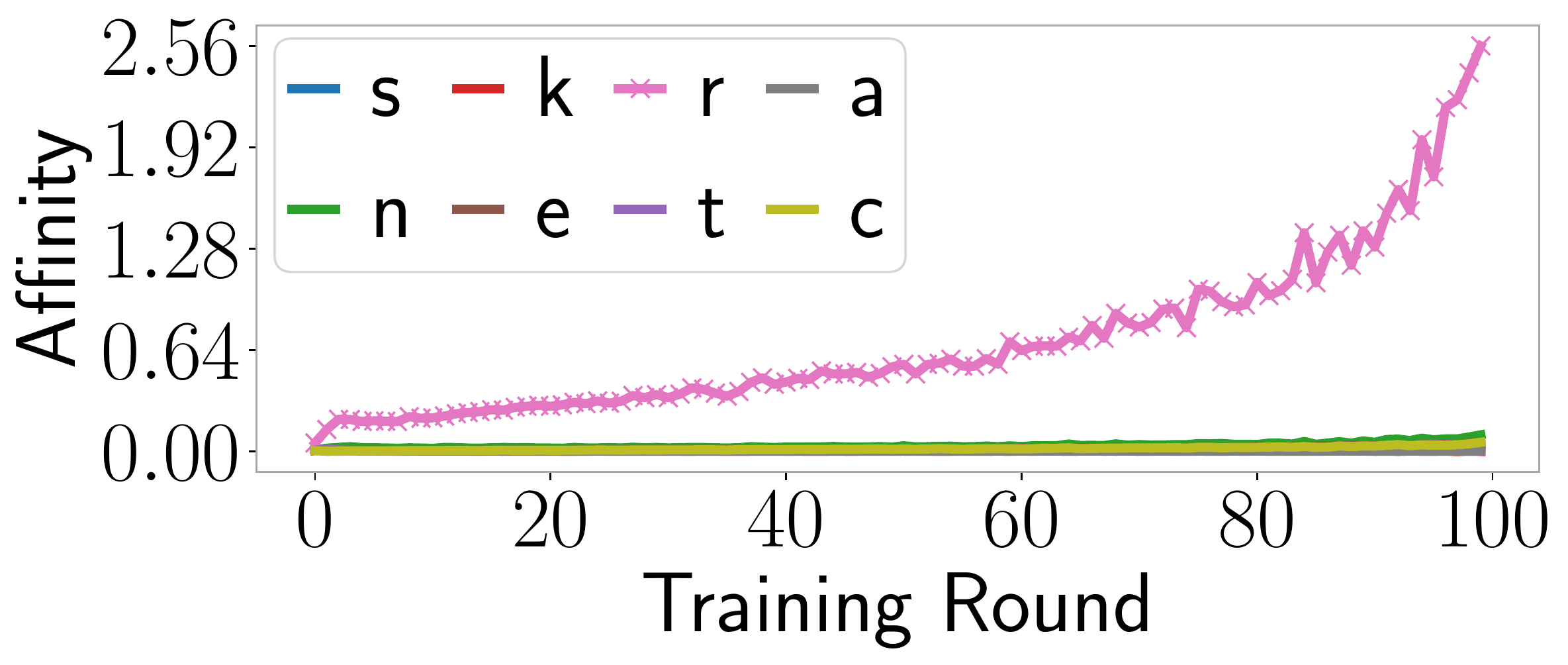}
    \caption{Affinities to activity \texttt{d}}
    \label{fig:affinity-to-d}
  \end{subfigure}
  \hfill    
  \begin{subfigure}[t]{0.31\textwidth}
    \includegraphics[width=\textwidth]{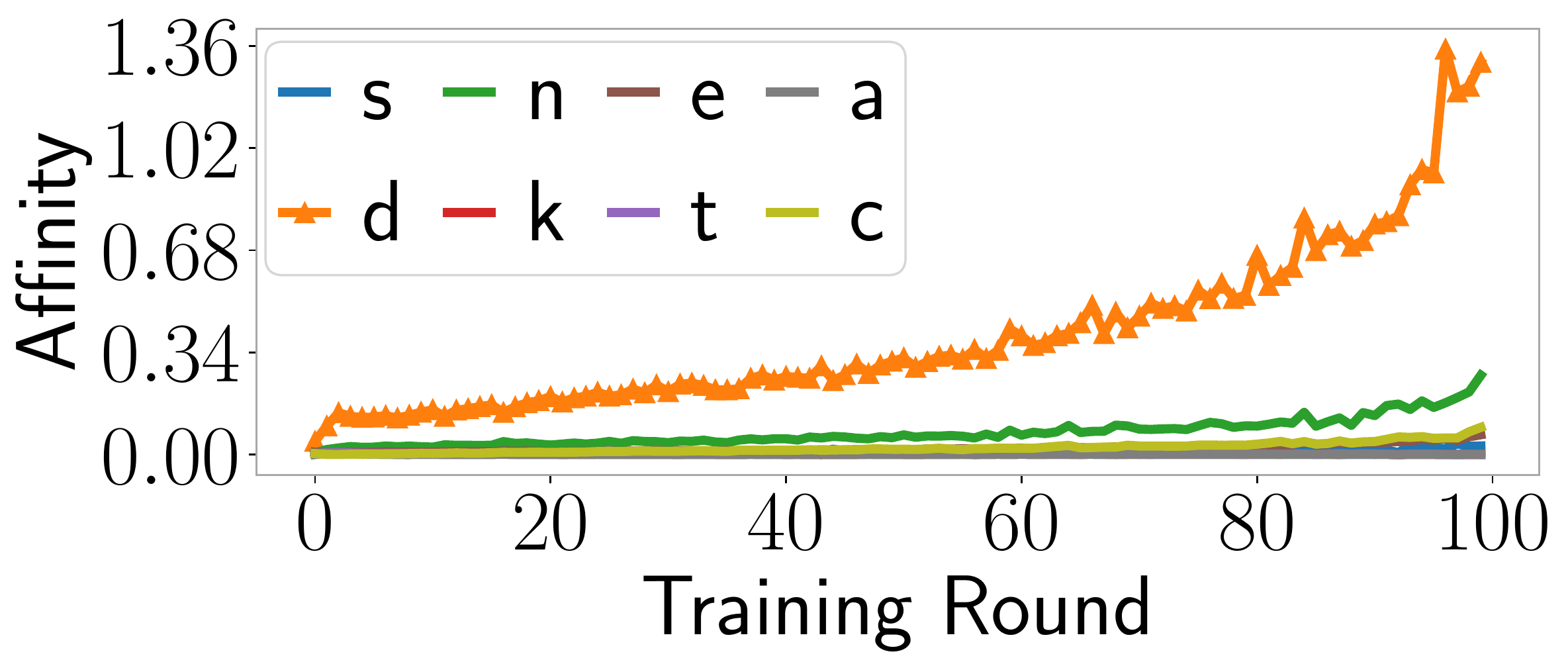}
    \caption{Affinities to activity \texttt{r}}
    \label{fig:affinity-to-r}
  \end{subfigure}  
  \hfill    
  \begin{subfigure}[t]{0.31\textwidth}
    \includegraphics[width=\textwidth]{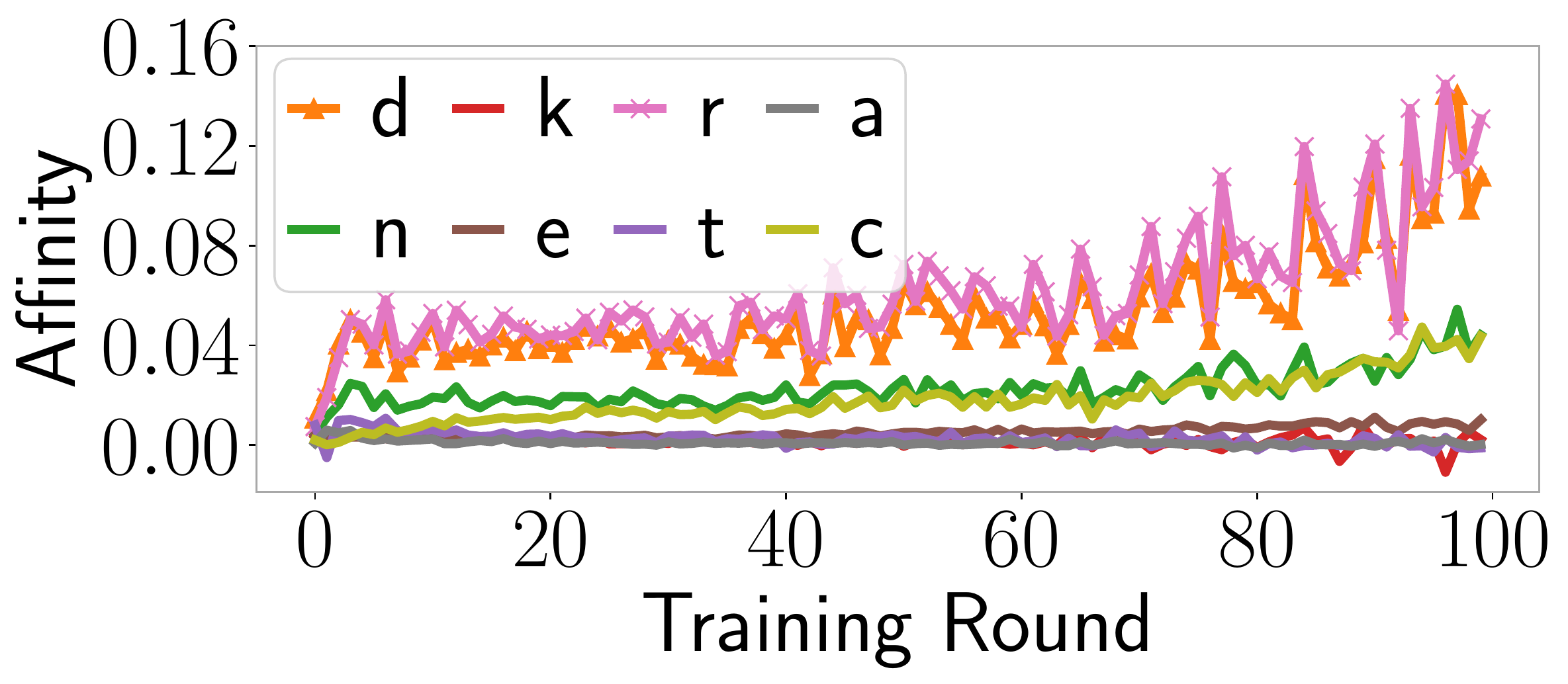}
    \caption{Affinities to activity \texttt{s}}
    \label{fig:affinity-to-s}
  \end{subfigure}    
 \caption{Changes of affinity scores of one activity to the other over the course of training on activity set \texttt{sdnkterca}. Activities \texttt{d} and \texttt{r} have high inter-activity affinity scores. The trends of affinities emerge at the early stage of training.}
 \label{fig:affinity-analysis}
\end{figure}


We demonstrate the effectiveness of our activity splitting approach by comparing it with the possible optimal and worst splits. The optimal and worst splits are obtained with two steps: 1) we measure the performance over all combinations of two splits and three splits of an activity set by training them from scratch;\footnote{There are fifteen and twenty-five combinations of two and three splits, respectively, for a set of five activities.} 2) we select the combination that yields the best performance as the optimal split and the worst performance as the worst split. 

Table \ref{tab:comparison-optimal-worst} compares the test loss of MuFL with the optimal and worst splits trained in two ways: 1) training each split from scratch; 2) training each split the same way as our activity splitting --- initializing models with the parameters obtained from all-in-one training. On the one hand,
training from initialization outperforms training from scratch in all settings. It suggests that initializing each split with all-in-one training model parameters can significantly improve the performance. On the other hand, our activity splitting method achieves the best performance in all settings, even though training from initialization reduces the gaps of different splits (the optimal and worst splits). These results indicate the effectiveness of our activity splitting approach.

\begin{figure}[t!]
  \centering
  \begin{subfigure}[t]{0.325\textwidth}
    \includegraphics[width=\textwidth]{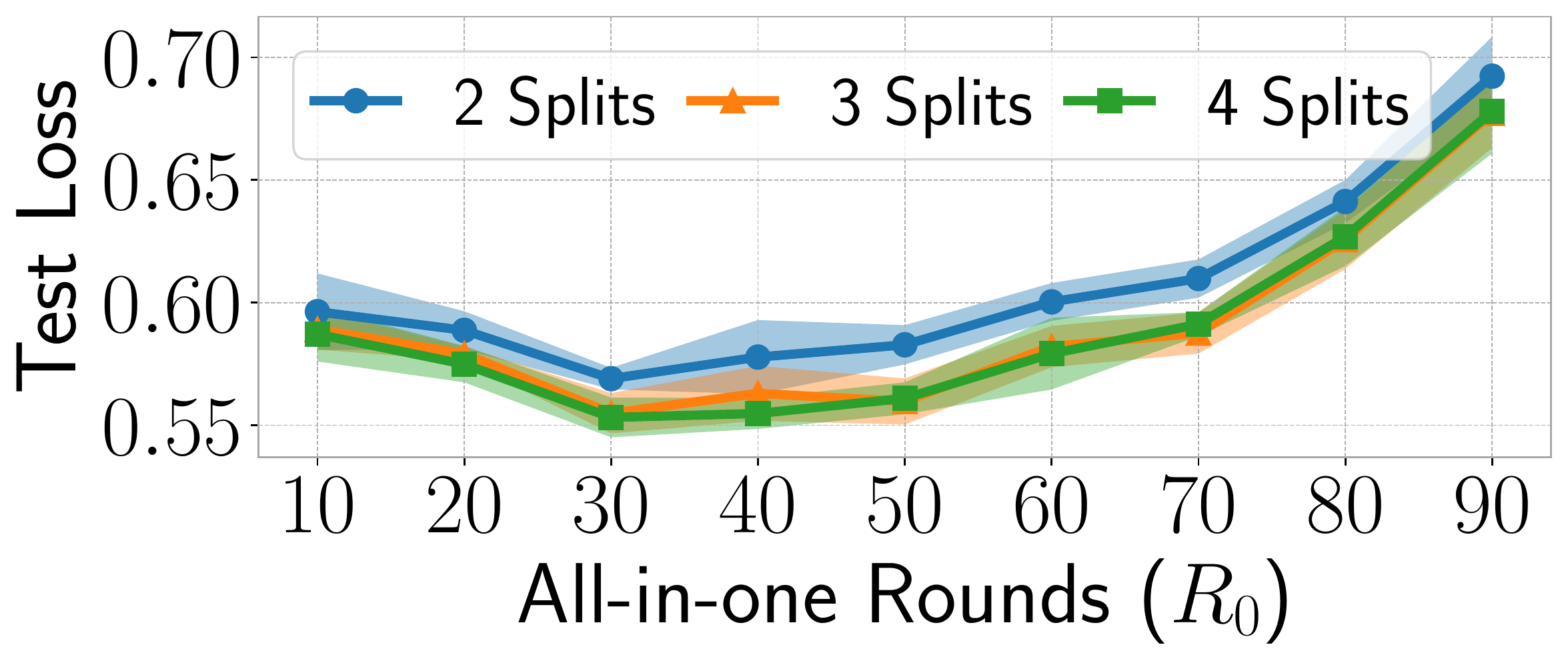}
    \caption{Training activity set: \texttt{sdnkt}}
    \label{fig:sdnkt-rounds}
  \end{subfigure}
  \hfill    
  \begin{subfigure}[t]{0.325\textwidth}
    \includegraphics[width=\textwidth]{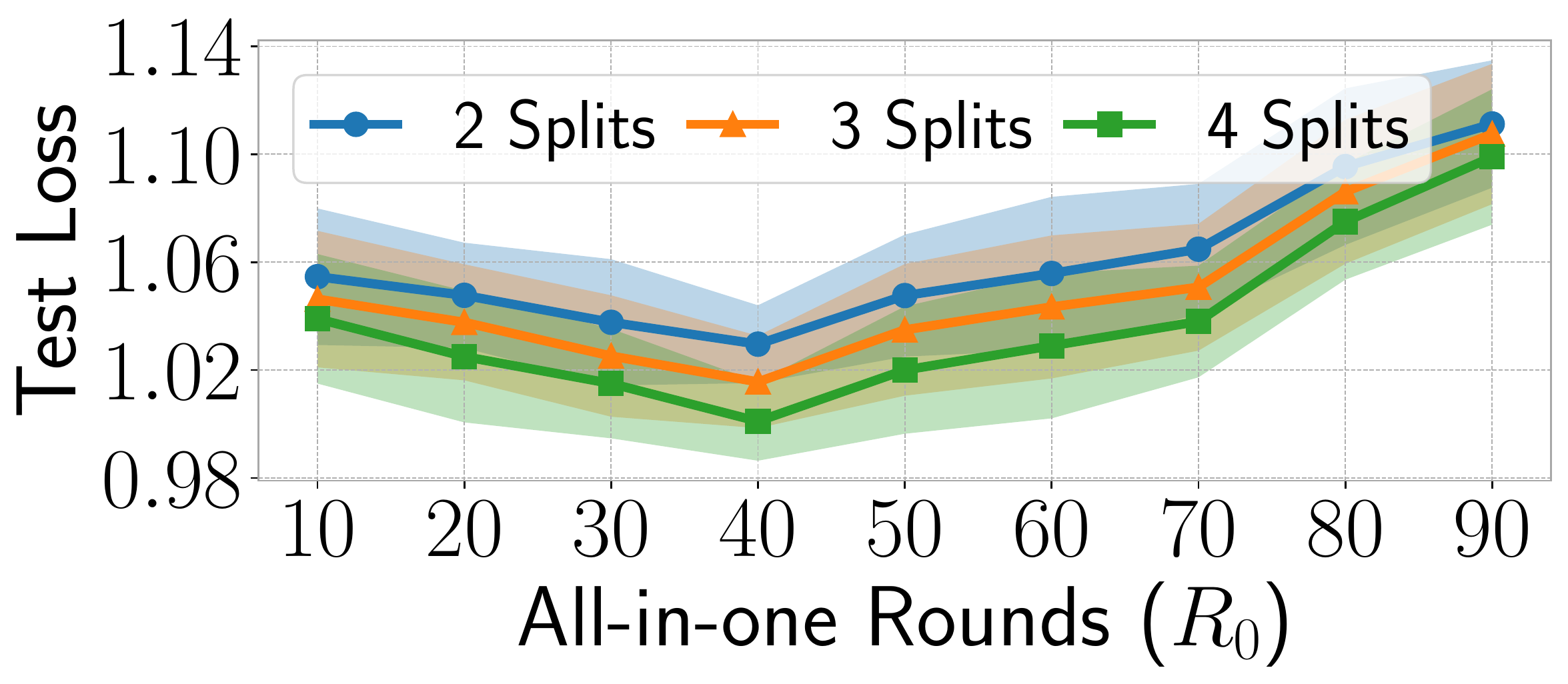}
    \caption{Training activity set: \texttt{erckt}}
    \label{fig:erckt-rounds}
  \end{subfigure}
  \hfill
  \begin{subfigure}[t]{0.325\textwidth}
    \includegraphics[width=\textwidth]{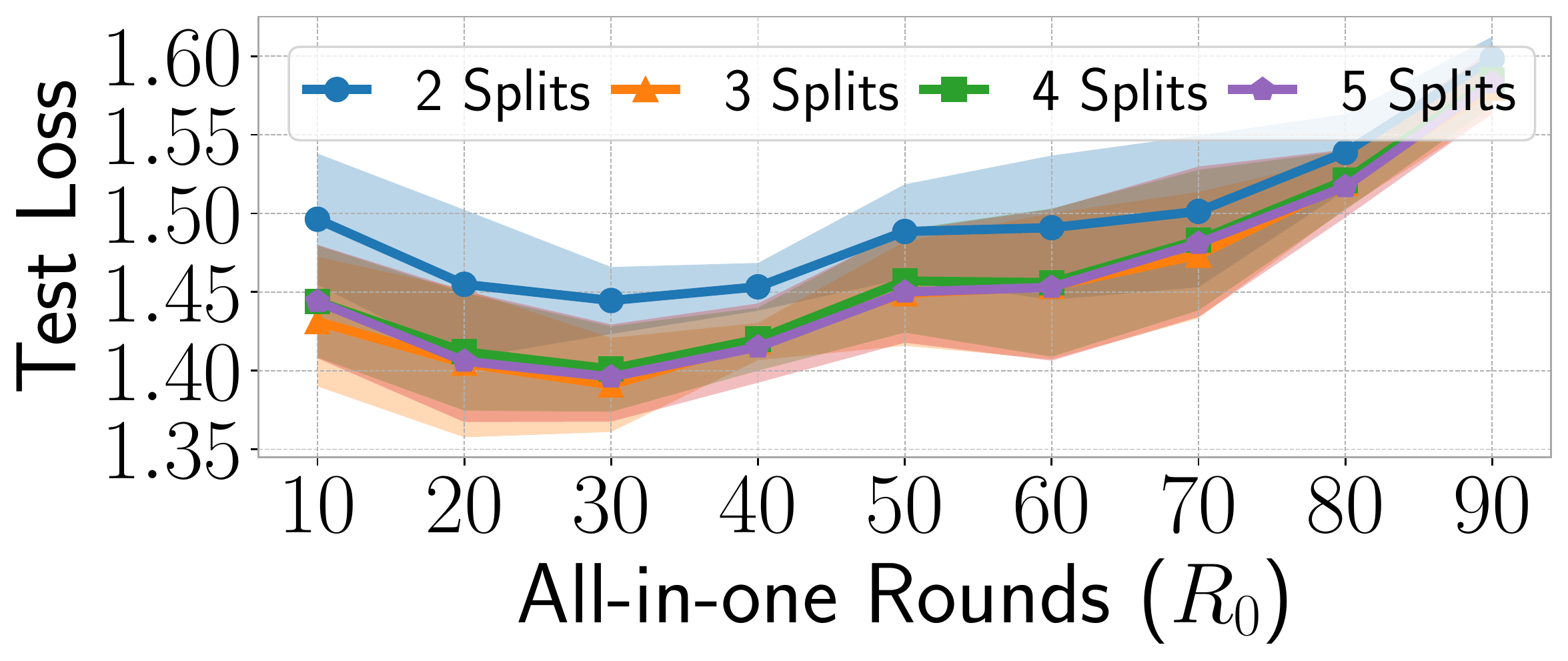}
    \caption{Training activity set: \texttt{sdnkterca}}
    \label{fig:sdnkterca-rounds}
  \end{subfigure}
 \caption{Performance comparison of training all-in-one activities for different $R_0$ rounds. Fixing the total training rounds $R = 100$, our method achieves the best performance when $R_0 = \{20, 30, 40\}$ rounds, varied over activity sets.}
 \label{fig:all-in-one-rounds}
\end{figure}

\begingroup
\setlength{\tabcolsep}{0.45em}
\begin{table}[t]\centering
  \caption{Performance of hierarchical splitting on three activity sets \texttt{sdnkt}, \texttt{erckt}, and \texttt{sdnkterca}. Hierarchical splitting outperforms two splits and achieves similar performance to three splits with less energy (kWh) consumption.}
  \label{tab:hierarchical-full}
  \begin{tabular}{cccccccccc}\toprule
  \multirow{2}{*}{Method} &\multicolumn{2}{c}{\texttt{sdnkt}} & &\multicolumn{2}{c}{\texttt{erckt} } & &\multicolumn{2}{c}{\texttt{sdnkterca}} \\\cmidrule{2-3}\cmidrule{5-6}\cmidrule{8-9}
  &Energy &Test Loss & &Energy &Test Loss & &Energy &Test Loss \\\midrule
  Two Splits &4.9 {\footnotesize ± 0.3} &0.578 {\footnotesize ± 0.015} & &6.7 {\footnotesize ± 0.2} &1.039 {\footnotesize ± 0.024} & &6.0 {\footnotesize ± 0.1} &1.445 {\footnotesize ± 0.021} \\
  Three Splits &5.4 {\footnotesize ± 0.7} &0.555 {\footnotesize ± 0.008} & &7.2 {\footnotesize ± 0.2} &1.015 {\footnotesize ± 0.018} & &6.6 {\footnotesize ± 0.4} &1.391 {\footnotesize ± 0.030} \\
  Hierarchical &\textbf{5.3 {\footnotesize ± 0.4}} &0.563 {\footnotesize ± 0.007} & &\textbf{6.9 {\footnotesize ± 0.2}} &1.022 {\footnotesize ± 0.020} & &\textbf{6.5 {\footnotesize ± 0.3}} &1.403 {\footnotesize ± 0.024} \\
  \bottomrule
  \end{tabular}
  \end{table}
\endgroup

\subsection{When to Split Training Activities?}

We further answer the question that how many $R_0$ rounds should we train the all-in-one activity before activity splitting. It is determined by two factors: 1) the rounds needed to obtain affinity scores for a reasonable activity splitting; 2) the rounds that yield the best overall performance.

\textbf{Affinity Analysis} \, We analyze changes in affinity scores over the course of training to show that early-stage affinity scores are acceptable for activity splitting. Figure \ref{fig:affinity-analysis} presents the affinity scores of different activities to one activity on activity set \texttt{sdnkterca}. Figure \ref{fig:affinity-to-d} and \ref{fig:affinity-to-r} indicate that activity \texttt{d} and activity \texttt{r} have high inter-activity affinity scores; they are divided into the same group as a result. In contrast, both \texttt{d} and \texttt{r} have high affinity score to activity \texttt{s} in Figure \ref{fig:affinity-to-s}, but not vice versa. These trends emerge in the early stage of training, thus, we employ the affinity scores of the \emph{tenth} round for activity splitting for the majority of experiments; they are effective in achieving promising results as shown in Figure \ref{fig:performance-vs-resource} and Table \ref{tab:comparison-optimal-worst}. We provide more affinity scores of other activities in Appendix \ref{apx:evaluation}. 



\textbf{The Impact of $R_0$ Rounds} \, Figure \ref{fig:all-in-one-rounds} compares the performance of training $R_0$ for 10 to 90 rounds before activity splitting on three activity sets. Fixing the total training round $R = 100$, we train each split of activities for $R_1 = R - R_0$ rounds. The results indicate that MuFL achieves the best performance when $R_0 = \{20, 30, 40\}$ rounds, varied over activity sets. Training the all-in-one activity for enough rounds helps utilize benefits and synergies of training together, but training for too many rounds almost suppresses the benefits of considering differences among activities. We suggest training $R_0$ for [20, 40] that strikes a good balance between these two extremes. 



\subsection{Hierarchical Splitting}

This section evaluates an alternative adaptive hierarchical activity splitting strategy. In activity splitting, we can divide the all-in-one training activity into $\{2, 3, \dots\}$ splits. As shown in Figure \ref{fig:performance-vs-resource}, more splits lead to better performance with slightly higher energy consumption in the five-activity set, but the trend is not straightforward in the nine-activity set. Apart from setting the number of splits directly, MuFL can split the training activity into more splits adaptively via two steps: 1) dividing the all-in-one activity into two splits and training each one for $R_1$ rounds; 2) further dividing one of them to two splits and train these three activities for $R_2$ rounds. 
We term it as \emph{hierarchical splitting} that adaptively divides activities into more splits. 

Table \ref{tab:hierarchical-full} compares the performance of hierarchical splitting (3 splits) with directly splitting to multiple splits on three activity sets \texttt{sdnkt}, \texttt{erckt}, and \texttt{sdnkterca}. We use $R_0 = 30$, $R_1 = 40$, and $R_2 = 30$ for  \texttt{sdnkt} and \texttt{erckt}, and $R_0 = 30$, $R_1 = 20$, and $R_2 = 50$ for \texttt{sdnkterca}. Hierarchical splitting effectively reduces test losses of two splits and achieves comparable performance to three splits with less energy consumption. These results suggest that hierarchical splitting can be an alternative method of activity splitting, which could be useful without determining the number of splits beforehand. Additionally, these results also demonstrate the possibility of other activity splitting strategies to be considered in future works. 

\subsection{Additional Ablation Studies}

\begin{wrapfigure}{r}{.5\textwidth}
  \begin{subfigure}[t]{0.31\textwidth}
    \includegraphics[width=\textwidth]{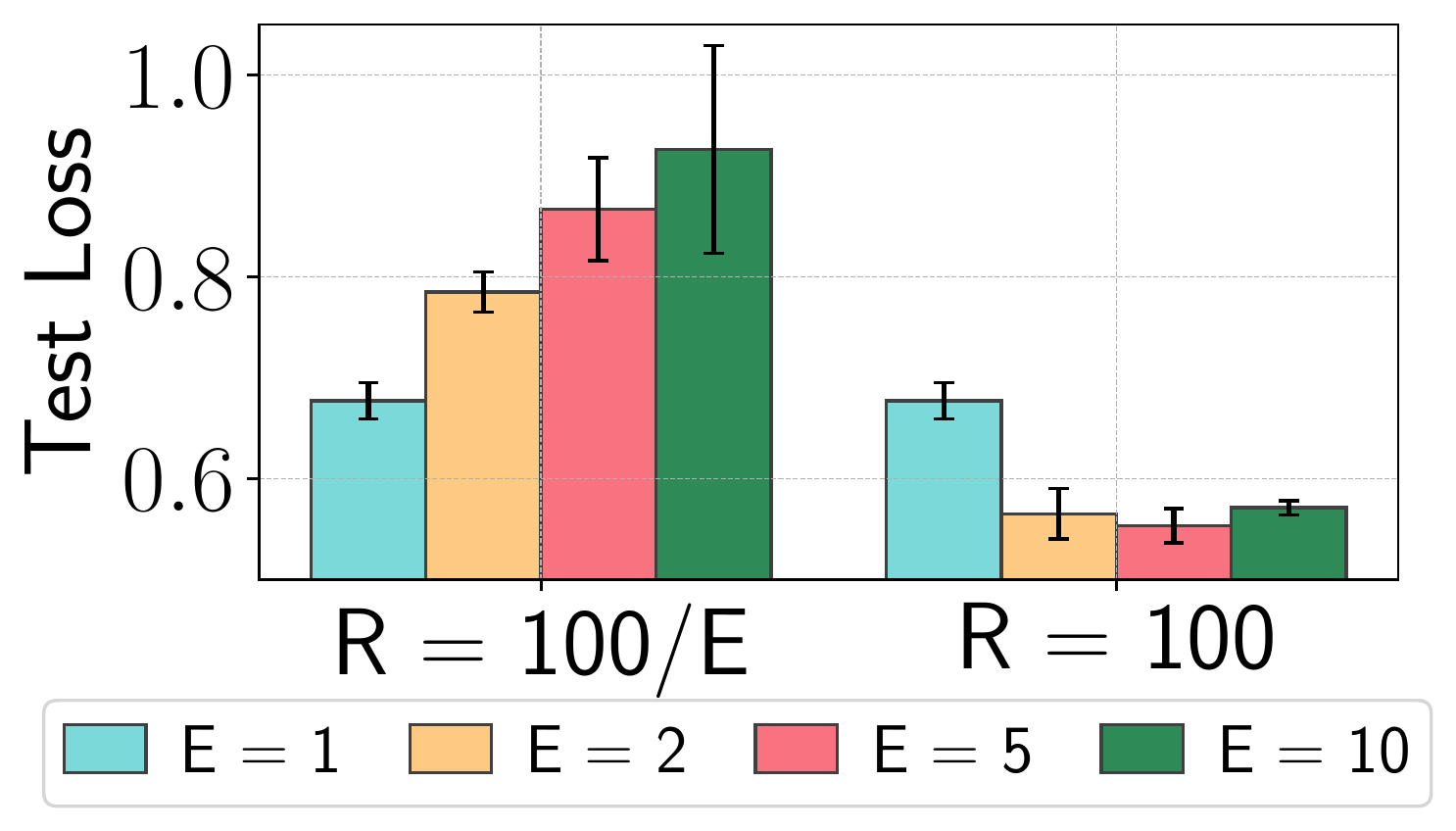}
    \caption{Impact of $E$}
    \label{fig:sdnkt-local-epoch}
  \end{subfigure}
  \begin{subfigure}[t]{0.18\textwidth}
    \includegraphics[width=\textwidth]{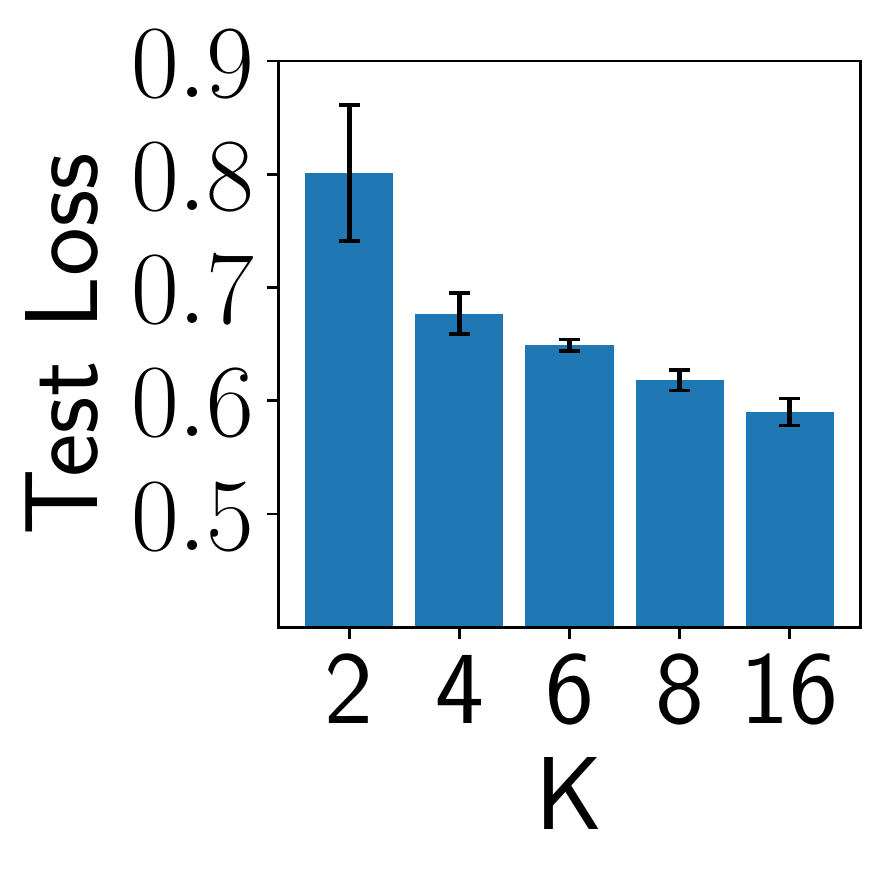}
    \caption{Impact of $K$}
    \label{fig:num-clients}
  \end{subfigure}

  \caption{Ablation studies on the impact of FL settings on activity set \texttt{sdnkt}: (a) local epoch $E$; (b) the number of selected clients $K$. Larger $E$ (with $R=100$) and $K$ requires higher computation. They could reduce losses, but the marginal benefit decreases as computation increases.}
  \label{fig:ablation}
\end{wrapfigure}

This section provides ablation studies of the impact of local epoch $E$ and the number of selected clients $K$ in FL using all-in-one training. We report the results of activity set \texttt{sdnkt} here and provide more results in Appendix \ref{apx:evaluation}.

\textbf{Impact of Local Epoch $E$} \, Local epoch defines the number of epochs each client trains before uploading training updates to the server. The total computation is $R * E$, where $R$ is the total training rounds. Figure ~\ref{fig:sdnkt-local-epoch} compares test losses of local epochs $E = \{1, 2, 5, 10\}$. On the one hand, 
fixing the total computation ($R = \nicefrac{100}{E}$), larger $E$ results in performance degradation. 
One the other hand, fixing training rounds $R = 100$, larger $E$ could lead to better performance, which is especially effective when $E$ increases from $E = 1$ to $E = 2$. However, such improvement is not consistant when $E = 10$. It suggests the limitation of simply increasing computation with larger $E$ in improving performance. Note that MuFL (Table \ref{tab:comparison-optimal-worst}) achieves better results than $E = 5$ with $\sim 5\times$ less computation.



\textbf{Impact of The Number of Selected Clients $K$} \, Figure \ref{fig:num-clients} compares test losses of the number of selected clients $K = \{2, 4, 6, 8, 16\}$ in each round. Increasing the number of selected clients improves the performance, but the effect becomes marginal as $K$ increases. Larger $K$ can also be considered as using more computation in each round. Similar to the results of the impact of $E$, simply increasing computation can only improve performance to a certain extent. It also shows the significance of MuFL that increases performance with slightly more computation. We use $K = 4$ by default for experiments and demonstrate that MuFL is also effective on $K = 8$ in Appendix \ref{apx:evaluation}.



\section{Conclusions and Discussions}
\label{sec:conclusion}

In this work, we propose a smart multi-tenant federated learning system to effectively coordinate and execute multiple simultaneous FL training activities. In particular, we introduce activity consolidation and activity splitting to consider both synergies and differences among training activities. Extensive empirical studies demonstrate that our method is effective in elevating performance and significant in reducing energy consumption and carbon footprint by more than 40\%, which are important metrics to our society. We believe that multi-tenant FL will emerge and empower many real-world applications with the fast development of FL. We hope this research will inspire the community to further work on algorithm and system optimizations of multi-tenant FL. Future work involves designing better scheduling mechanisms to coordinate training activities, employing and integrating client selection strategies to optimize resource and training allocation, and extending our optimization approaches to other multi-tenant FL scenarios. Lastly, studies of FL are closely related to data privacy risks and the fairness among training activities could also be considered; the applications of multi-tenant FL should seek tools and new approaches to address these issues.

\bibliography{neurips_2022}
\bibliographystyle{plain}

\newpage


\appendix

The following appendixes provide supplemental material for the main manuscript. We update two parts compared to the one appended at the end of the main paper: 1) we update three runs of experiment results in Figure \ref{fig:erckt-local-epoch} and Figure \ref{fig:sdnkterca-local-epoch}; 2) we provide results of standalone training that conducts training using data in each client independently in Figure \ref{fig:standalone}.

\section{Multi-tenant FL Scenarios}
\label{apx:scenarios}

\begin{figure}[h]
  \begin{center}
  \includegraphics[width=0.9\linewidth]{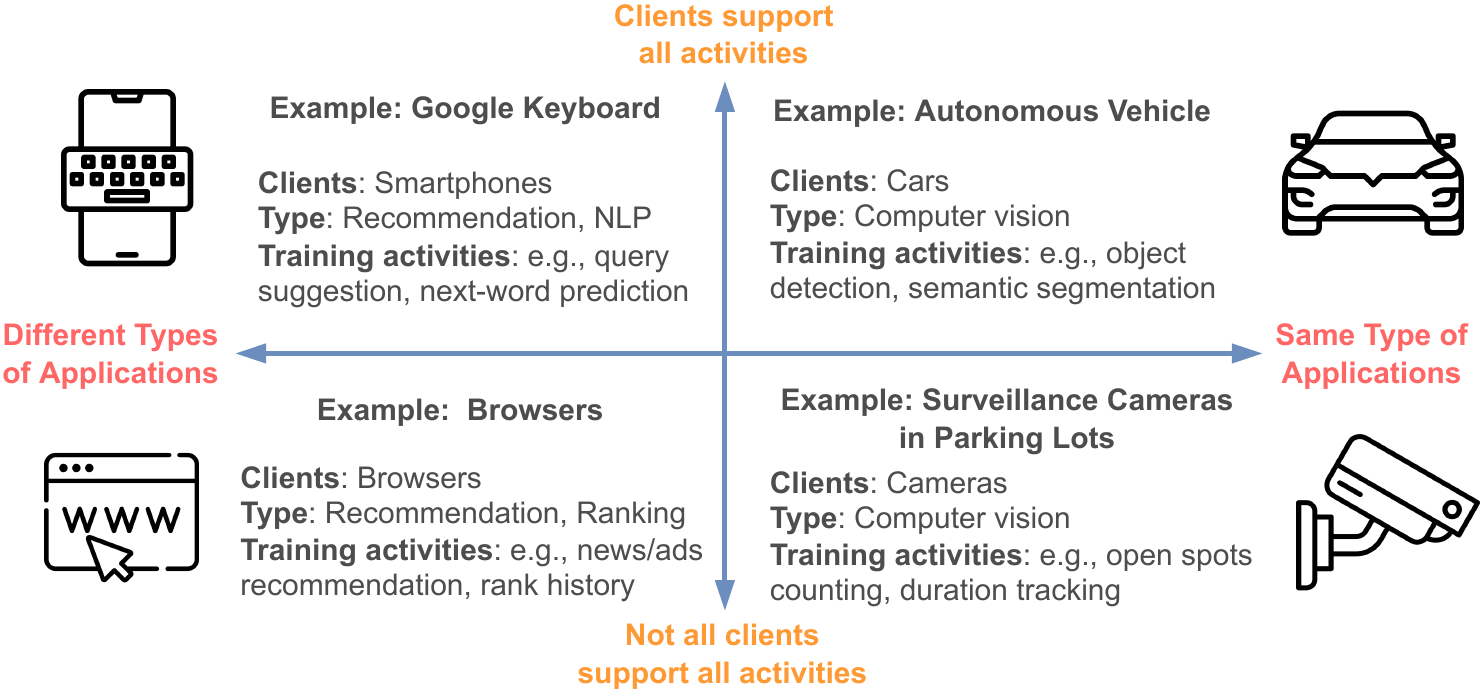}
     \caption{Illustration of the four multi-tenant FL scenarios.}
  \label{fig:scenarios}
  \end{center}
\end{figure}

We introduce four multi-tenant federated learning scenarios in Section \ref{sec:problem-setup}. Figure \ref{fig:scenarios} depicts these four scenarios with variances in two aspects: 1) whether all training activities are the same type of application, e.g., CV applications; 2) whether all clients support all training activities.

\section{Experimental Details}
\label{apx:experimental-setup}

This section provides more experimental information, including dataset, implementation details, and computation resources used.


\begin{wrapfigure}{r}{.4\textwidth}
  \begin{center}
  \includegraphics[width=\linewidth]{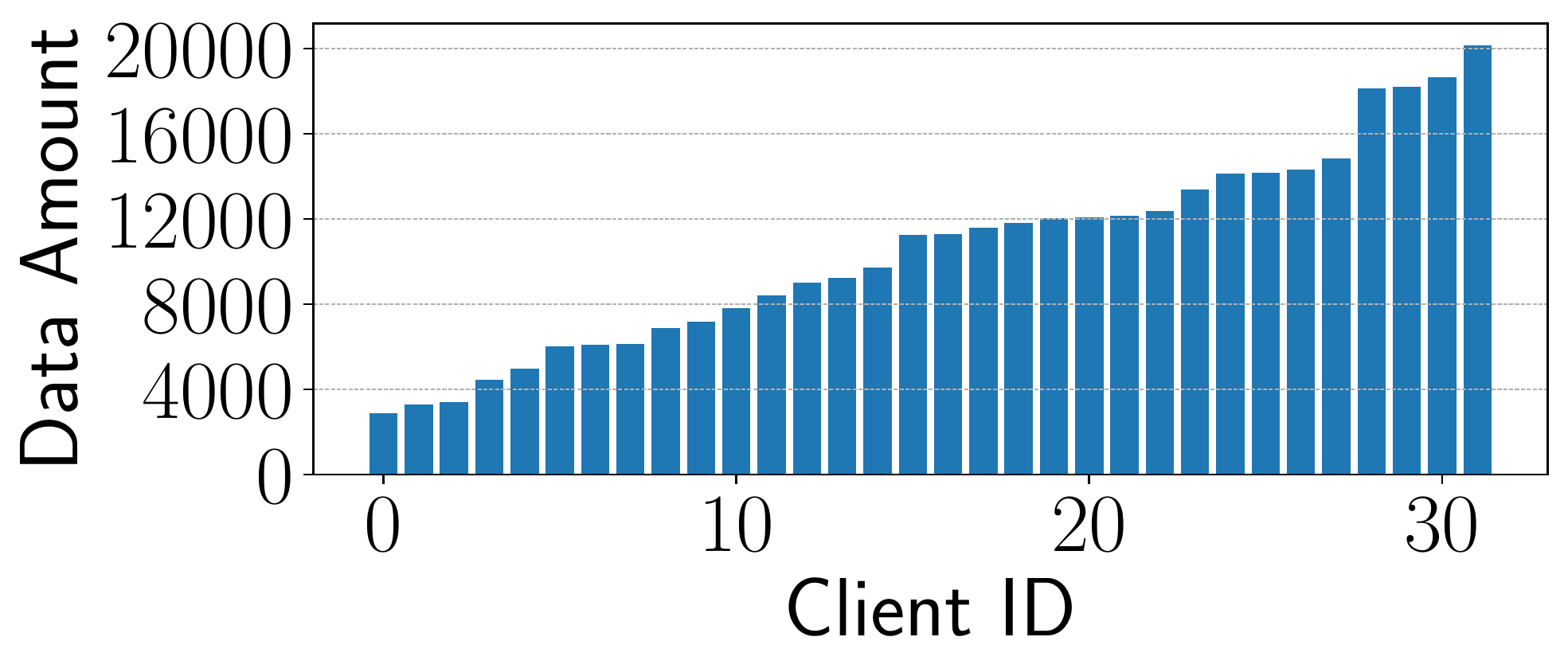}
     \caption{The data amount distribution of each training activity over 32 clients.}
  \label{fig:client-stats}
  \end{center}
\end{wrapfigure}

\begin{figure}[t!]
  \captionsetup[sub]{font=scriptsize}
  \centering
  \begin{subfigure}[t]{0.19\textwidth}
    \includegraphics[width=\textwidth]{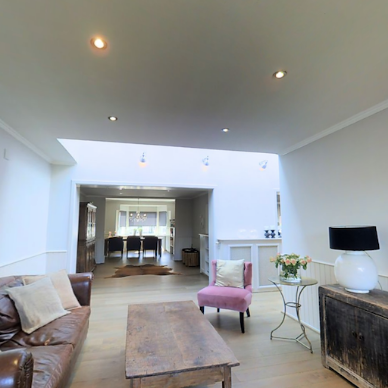}
    \caption{Input Image}
  \end{subfigure}
  \begin{subfigure}[t]{0.19\textwidth}
    \includegraphics[width=\textwidth]{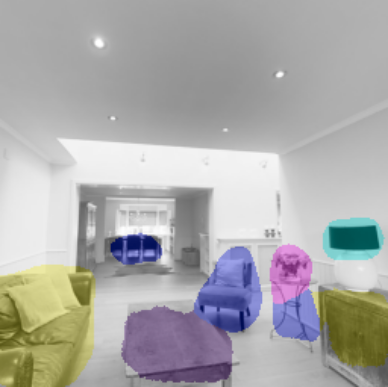}
    \caption{\texttt{s}: Segmentation}
  \end{subfigure}
  \begin{subfigure}[t]{0.19\textwidth}
    \includegraphics[width=\textwidth]{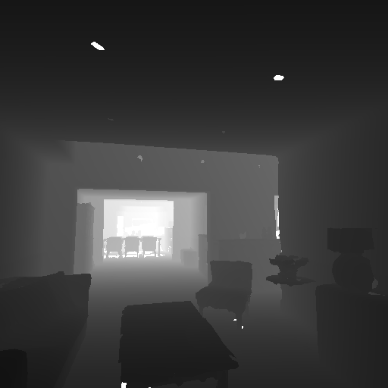}
    \caption{\texttt{d}: Depth Estimation}
  \end{subfigure}
  \begin{subfigure}[t]{0.19\textwidth}
    \includegraphics[width=\textwidth]{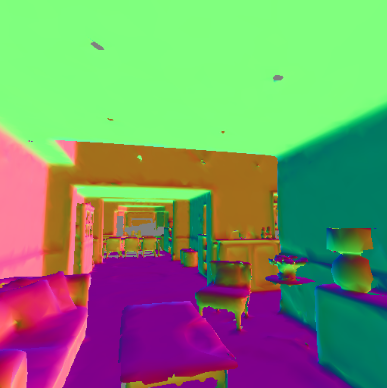}
    \caption{\texttt{n}: Normal}
  \end{subfigure}
  \begin{subfigure}[t]{0.19\textwidth}
    \includegraphics[width=\textwidth]{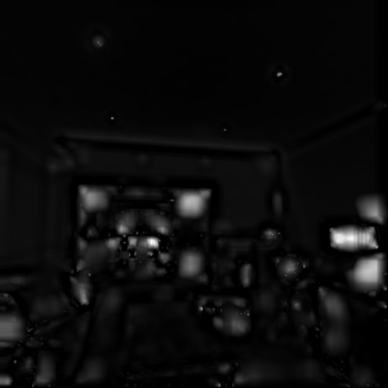}
    \caption{\texttt{k}: Keypoint}
  \end{subfigure}
  \begin{subfigure}[t]{0.19\textwidth}
    \includegraphics[width=\textwidth]{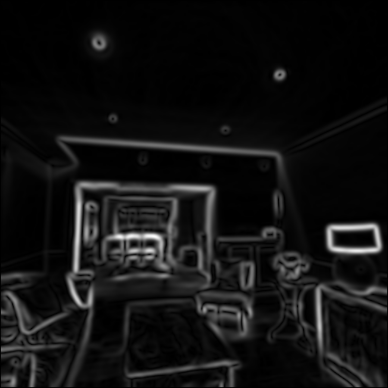}
    \caption{\texttt{t}: Edge Texture}
  \end{subfigure}
  \begin{subfigure}[t]{0.19\textwidth}
    \includegraphics[width=\textwidth]{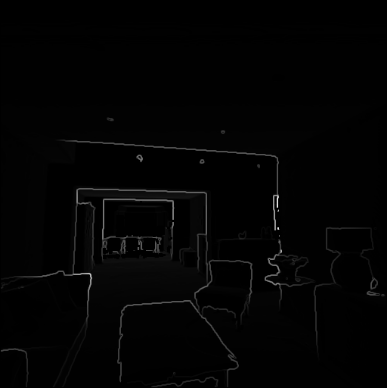}
    \caption{\texttt{e}: Edge Occlusion}
  \end{subfigure}
  \begin{subfigure}[t]{0.19\textwidth}
    \includegraphics[width=\textwidth]{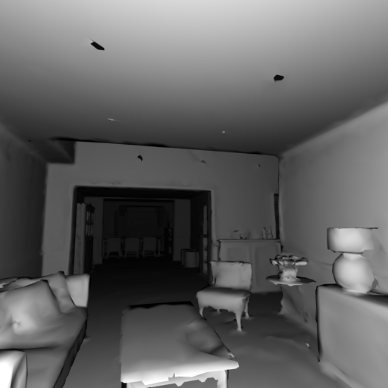}
    \caption{\texttt{r}: Reshasing}
  \end{subfigure}  
  \begin{subfigure}[t]{0.19\textwidth}
    \includegraphics[width=\textwidth]{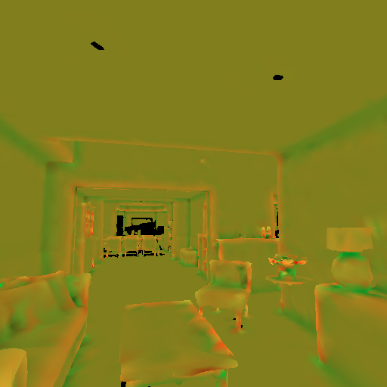}
    \caption{\texttt{c}: Principle Curvature}
  \end{subfigure}
  \begin{subfigure}[t]{0.19\textwidth}
    \includegraphics[width=\textwidth]{figures/activities/auto_encoder.png}
    \caption{\texttt{a}: Auto-encoder}
  \end{subfigure}  
 \caption{Sample images of nine training activities corresponding to the input image.}
 \label{fig:sample-activities}
\end{figure}

\begin{figure}[t!]
  \centering
  \begin{subfigure}[t]{0.19\textwidth}
    \includegraphics[width=\textwidth]{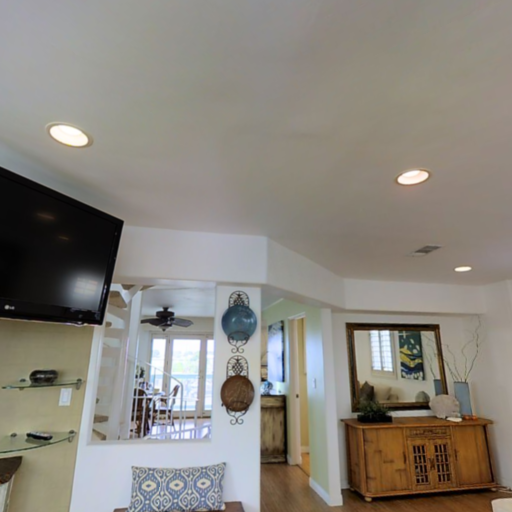}
  \end{subfigure}
  \begin{subfigure}[t]{0.19\textwidth}
    \includegraphics[width=\textwidth]{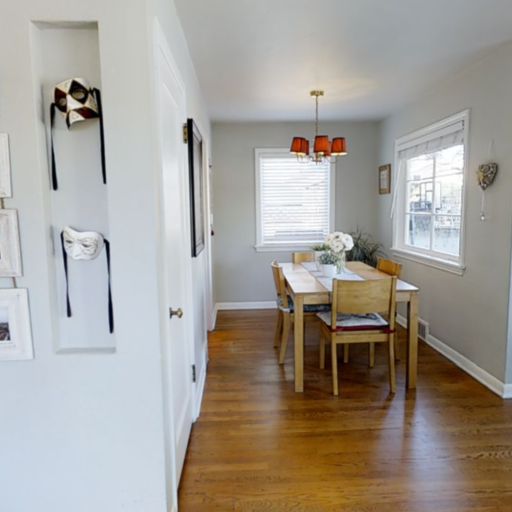}
  \end{subfigure}
  \begin{subfigure}[t]{0.19\textwidth}
    \includegraphics[width=\textwidth]{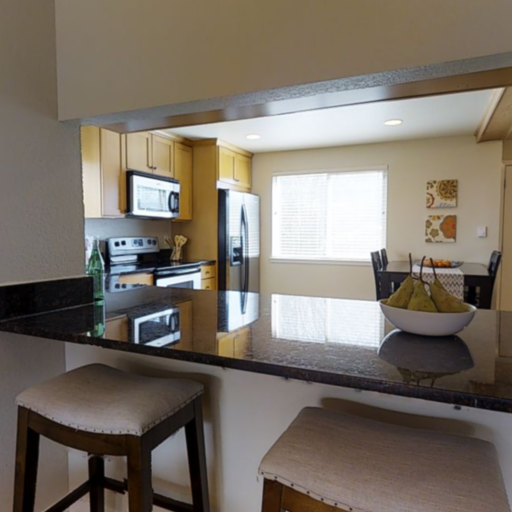}
  \end{subfigure}
  \begin{subfigure}[t]{0.19\textwidth}
    \includegraphics[width=\textwidth]{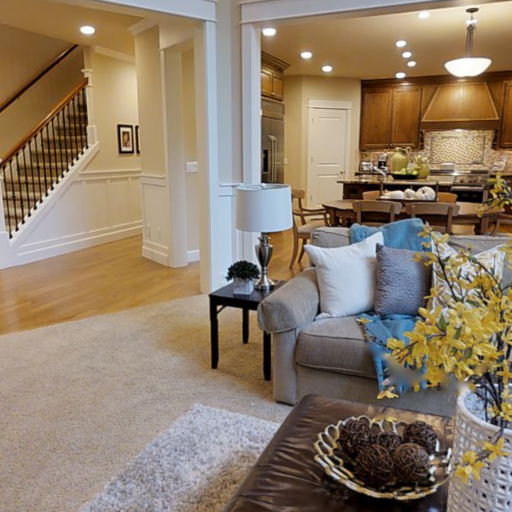}
  \end{subfigure}
  \begin{subfigure}[t]{0.19\textwidth}
    \includegraphics[width=\textwidth]{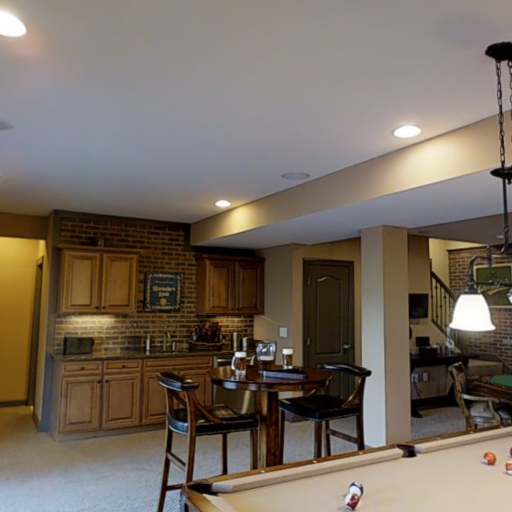}
  \end{subfigure}
 \caption{Sample images of five clients, where each client contains indoor scenes of a building. These indoor images differ in design, layout, objects, and illumination.}
 \label{fig:sample-clients}
\end{figure}

\paragraph{Dataset} We run experiments using Taskonomy dataset \cite{zamir2018taskonomy}, which is a large computer vision (CV) dataset of indoor scenes of buildings. To facilitate reproducibility and mitigate computational requirements, we use the tiny split of Taskonomy dataset,\footnote{Taskonomy dataset is released under MIT license and can be downloaded from their official repository 
\url{https://github.com/StanfordVL/taskonomy}.} whose size is around 445GB. We select nine CV applications to form three sets of training activities: \texttt{sdnkt}, \texttt{erckt}, \texttt{sdnkterca}. These nine actvities are also used in \cite{standley2020which}. Figure \ref{fig:sample-activities} provides sample images of these nine training activities, as well as the representation of each character.\footnote{The meaning of each character in \texttt{sdnkterca} are as follows; \texttt{s}: semantic segmentation, \texttt{d}: depth estimation, \texttt{n}: normals, \texttt{k}: keypoint, \texttt{t}: edge texture, \texttt{e}: edge occlusion, \texttt{r}: reshaping, \texttt{c}: principle curvature, \texttt{a}: auto-encoder.}  In particular, we employ indoor images of 32 buildings \footnote{The name of the buildings are allensville, beechwood, benevolence, coffeen, collierville, corozal, cosmos, darden, forkland, hanson, hiteman, ihlen, klickitat, lakeville, leonardo, lindenwood, markleeville, marstons, mcdade, merom, mifflinburg, muleshoe, newfields, noxapater, onaga, pinesdale, pomaria, ranchester, shelbyville, stockman, tolstoy, and uvalda.} as the total number of clients $N = 32$; each client contains images of a building to simulate the statistical heterogeneity. On the one hand, clients have different sizes of data. Figure \ref{fig:client-stats} shows the distribution of dataset sizes of an activity of clients. On the other hand, Figure \ref{fig:sample-clients} shows sample images of five clients; their indoor scenes vary in design, layout, objects, and illumination. 


\paragraph{Implementation Details} We implement multi-tenant FL systems in Python using EasyFL \cite{zhuang2022easyfl} and PyTorch \cite{paszke2017pytorch}. We simulate the FL training on a cluster of NVIDIA Tesla V100 GPUs, where each node in the cluster contains 8 GPUs. In each round, each selected client is allocated to a GPU to conduct training; these clients communicate via the NCCL backend. Besides, we employ FedAvg \cite{fedavg} for the server aggregation. By default, we randomly select $K = 4$ clients to train for $E = 1$ local epochs in each round and train for $R = 100$ rounds.

We reference the implementation of multi-task learning from \cite{standley2020which}'s official repository \footnote{\url{https://github.com/tstandley/taskgrouping}} for all-in-one training and training of each split after activity splitting. Particularly, the network architecture contains an encoder $\theta_s$ and multiple decoders $\theta_{\alpha_i}$; one decoder for a training activity $\alpha_i$. We use the modified Xception Network \cite{chollet2017xception} as the encoder for activity sets \texttt{sdnkt} and \texttt{erckt} and half size of the network (half amount of parameters) for activity set \texttt{sdnkterca}. The decoders contain four deconvolution layers and four convolution layers. The batch size is $B = 64$ for \texttt{sdnkt} and \texttt{erckt} and $B = 32$ for \texttt{sdnkterca}. These are the maximum batch sizes for one GPU without out-of-memory issues. In addition, we use polynomial learning rate decay $(1 - \frac{r}{R})^{0.9}$ to update learning rate in each round with initial learning rate $\eta = 0.1$, where $r$ is the number of trained rounds and $R = 100$ is the default total training rounds. The optimizer is stochastic gradient descent (SGD), with momentum of 0.9 and weight decay $1e^{-4}$. 

\paragraph{Implementation of Compared Methods} GradNorm \cite{chen2018gradnorm} implementation is adopted from \cite{standley2020which,fifty2021tag} with default $\alpha=1.5$ and TAG \cite{fifty2021tag} implementation is adopted from their official repository \footnote{\url{https://github.com/google-research/google-research/tree/master/tag}}. Next, we provide the details of how we compute results of HOA \cite{standley2020which} and TAG \cite{fifty2021tag}. 

HOA \cite{standley2020which} needs to compute test losses for individual activities and pair-wise activity combinations for $R = 100$ rounds. After that, we use these results to estimate test losses of higher-order combinations following \cite{standley2020which}. We then compute the actual test losses for the optimal activity splits that have the lowest test losses by training them from scratch. For example, for activity set \texttt{sdnkt}, we compute \texttt{s}, \texttt{d}, \texttt{n}, \texttt{k}, \texttt{t} and ten pair-wise activity combinations. Then, we use these results to estimate test losses of higher-order combinations.

TAG \cite{fifty2021tag} first computes all-in-one training for $R=100$ rounds to obtain the pair-wise affinities. Then, it uses a network selection algorithm to group these activities. After that, we train each group of activities from scratch for $R = 100$ rounds to obtain test losses. The best result is reported for overlapping activities. For example, \{\texttt{sd, dn, kt}\} is the best result of three splits of TAG on activity set \texttt{sdnkt}. Then, each split is trained from scratch to obtain test losses.

\paragraph{Computation Resources} Experiments in this work take approximately 27,765 GPU hours of NVIDIA Tesla V100 GPU for training. We conduct three independent runs of experiments for the majority of empirical studies. In each run, activity set \texttt{sdnkt} takes around 2,330 GPU hours, \texttt{erckt} takes around 3,280 GPU hours, and \texttt{sdnkterca} takes around 3,645 GPU hours. These include experiments of compared methods and ablation studies, whereas these do not include the GPU hours for validation and testing. It takes around the same GPU hours as training when we validate the model after each training round.

\begingroup
\setlength{\tabcolsep}{0.2em}
\begin{table}[t]\centering
  \caption{Comparison of test loss, energy consumption, and carbon footprint on activity set \texttt{sdnkt}.}\label{tab:sdnkt-full}
  \scriptsize
  \begin{tabular}{c|c|c|c|c|c|c|c|c|c}\toprule
    Method &Splits &Energy (kWh) &CO2eq (g) &Total Loss &s &d &n &k &t \\\midrule
    One by one &- &8.4 ± 0.1 &2465 ± 39 &0.603 ± 0.030 &0.086 ± 0.005 &0.261 ± 0.023 &0.107 ± 0.001 &0.107 ± 0.003 &0.043 ± 0.002 \\
    All-in-one &- &3.7 ± 0.1 &1086 ± 28 &0.677 ± 0.018 &0.087 ± 0.002 &0.246 ± 0.010 &0.136 ± 0.001 &0.126 ± 0.019 &0.083 ± 0.008 \\
    GradNorm &- &4.1 ± 0.4 &1200 ± 122 &0.691 ± 0.013 &0.092 ± 0.001 &0.251 ± 0.012 &0.138 ± 0.003 &0.118 ± 0.007 &0.093 ± 0.019 \\\midrule
    HOA &2 &31.0 ± 0.5 &9125 ± 140 &0.651 ± 0.029 &0.091 ± 0.011 &0.245 ± 0.002 &0.135 ± 0.000 &0.107 ± 0.003 &0.074 ± 0.023 \\
    TAG &2 &9.8 ± 0.3 &2876 ± 88 &0.624 ± 0.015 &0.083 ± 0.004 &0.242 ± 0.005 &0.134 ± 0.001 &0.110 ± 0.007 &0.055 ± 0.006 \\
    \textbf{MuFL} &2 &4.9 ± 0.3 &1431 ± 94 &0.578 ± 0.015 &0.069 ± 0.006 &0.231 ± 0.006 &0.124 ± 0.002 &0.102 ± 0.003 &0.052 ± 0.003 \\\midrule
    HOA &3 &31.0 ± 0.5 &9125 ± 140 &0.598 ± 0.029 &0.083 ± 0.022 &0.239 ± 0.007 &0.127 ± 0.008 &0.107 ± 0.003 &0.043 ± 0.002 \\
    TAG &3 &11.3 ± 0.2 &3313 ± 56 &0.613 ± 0.032 &0.094 ± 0.005 &0.233 ± 0.002 &0.122 ± 0.013 &0.110 ± 0.008 &0.055 ± 0.008 \\
    \textbf{MuFL} &3 &5.4 ± 0.3 &1589 ± 94 &0.555 ± 0.015 &0.072 ± 0.006 &0.222 ± 0.006 &0.124 ± 0.002 &0.095 ± 0.003 &0.042 ± 0.003 \\\midrule
    HOA &4 &31.0 ± 0.5 &9125 ± 140 &0.597 ± 0.015 &0.094 ± 0.009 &0.238 ± 0.002 &0.115 ± 0.014 &0.107 ± 0.003 &0.043 ± 0.002 \\
    TAG &4 &13.7 ± 0.3 &4016 ± 80 &0.603 ± 0.027 &0.083 ± 0.005 &0.233 ± 0.002 &0.122 ± 0.013 &0.110 ± 0.008 &0.055 ± 0.008 \\
    \textbf{MuFL} &4 &6.7 ± 0.3 &1969 ± 75 &0.548 ± 0.001 &0.070 ± 0.002 &0.230 ± 0.008 &0.111 ± 0.000 &0.095 ± 0.007 &0.042 ± 0.001 \\
    \bottomrule
  \end{tabular}
  \end{table}
\endgroup

\begingroup
\setlength{\tabcolsep}{0.2em}
\begin{table}[t]\centering
  \caption{Comparison of test loss, energy consumption, and carbon footprint on activity set \texttt{erckt}.}\label{tab:erckt-full}
  \scriptsize
  \begin{tabular}{c|c|c|c|c|c|c|c|c|c}\toprule
  Method &Splits &Energy (kWh) &CO2eq (g) &Total Loss &e &r &c &k &t \\\midrule
  One by one &- &11.1 ± 2.2 &3277 ± 660 &1.055 ± 0.034 &0.148 ± 0.000 &0.371 ± 0.029 &0.386 ± 0.006 &0.107 ± 0.003 &0.043 ± 0.002 \\
  All-in-one &- &5.0 ± 0.3 &1478 ± 84 &1.130 ± 0.022 &0.146 ± 0.001 &0.379 ± 0.019 &0.393 ± 0.002 &0.110 ± 0.003 &0.079 ± 0.013 \\
  GradNorm &- &5.0 ± 0.2 &1462 ± 70 &1.154 ± 0.055 &0.147 ± 0.002 &0.381 ± 0.015 &0.394 ± 0.001 &0.149 ± 0.062 &0.082 ± 0.005 \\\midrule
  HOA &2 &38.3 ± 0.3 &11265 ± 86 &1.082 ± 0.032 &0.149 ± 0.003 &0.365 ± 0.025 &0.394 ± 0.002 &0.109 ± 0.002 &0.064 ± 0.022 \\
  TAG &2 &14.0 ± 0.9 &4119 ± 279 &1.095 ± 0.033 &0.147 ± 0.002 &0.379 ± 0.013 &0.393 ± 0.000 &0.108 ± 0.005 &0.068 ± 0.015 \\
  \textbf{MuFL} &2 &6.7 ± 0.2 &1957 ± 53 &1.039 ± 0.024 &0.143 ± 0.001 &0.343 ± 0.014 &0.393 ± 0.001 &0.104 ± 0.006 &0.056 ± 0.007 \\\midrule
  HOA &3 &38.3 ± 0.2 &11265 ± 53 &1.062 ± 0.024 &0.149 ± 0.001 &0.365 ± 0.014 &0.394 ± 0.001 &0.109 ± 0.006 &0.046 ± 0.007 \\
  TAG &3 &14.4 ± 0.6 &4242 ± 170 &1.091 ± 0.034 &0.147 ± 0.002 &0.388 ± 0.014 &0.396 ± 0.002 &0.109 ± 0.009 &0.050 ± 0.011 \\
  \textbf{MuFL} &3 &7.2 ± 0.2 &2108 ± 50 &1.015 ± 0.018 &0.143 ± 0.000 &0.336 ± 0.005 &0.383 ± 0.001 &0.102 ± 0.008 &0.052 ± 0.009 \\\midrule
  HOA &4 &38.3 ± 0.3 &11265 ± 86 &1.053 ± 0.034 &0.148 ± 0.002 &0.369 ± 0.028 &0.386 ± 0.006 &0.105 ± 0.001 &0.045 ± 0.003 \\
  TAG &4 &17.4 ± 0.5 &5114 ± 159 &1.087 ± 0.028 &0.147 ± 0.002 &0.384 ± 0.011 &0.396 ± 0.002 &0.109 ± 0.009 &0.050 ± 0.011 \\
  \textbf{MuFL} &4 &7.6 ± 0.0 &2229 ± 14 &1.002 ± 0.014 &0.143 ± 0.000 &0.336 ± 0.005 &0.383 ± 0.001 &0.094 ± 0.009 &0.046 ± 0.004 \\
  \bottomrule
  \end{tabular}
\end{table} 
\endgroup

\begingroup
\setlength{\tabcolsep}{0.04em}
\begin{table}[t]\centering
  \caption{Comparison of test loss, energy consumption, and carbon footprint on \texttt{sdnkterca}.}\label{tab:sdnkterca-full}
  \tiny
  \begin{tabular}{c|c|c|c|c|c|c|c|c|c|c|c|c|c}\toprule
    Method &Splits &Energy &CO2eq (g) &Total Loss &s &d & n &k &t &e &r &c &a \\\midrule
    One by one &- &11.9 ±0.5 &3512 ±151 &1.46 ±0.011 &0.08 ±0.009 &0.24 ±0.014 &0.10 ±0.001 &0.10 ±0.002 &0.04 ±0.003 &0.15 ±0.001 &0.35 ±0.011 &0.38 ±0.002 &0.02 ±0.000 \\
    All-in-one &- &4.9 ±0.2 &1435 ±60 &1.49 ±0.025 &0.09 ±0.002 &0.23 ±0.009 &0.13 ±0.002 &0.10 ±0.002 &0.07 ±0.005 &0.14 ±0.001 &0.33 ±0.011 &0.39 ±0.001 &0.02 ±0.001 \\
    GradNorm &- &5.3 ±1.3 &1561 ±377 &1.50 ±0.049 &0.08 ±0.004 &0.24 ±0.014 &0.13 ±0.003 &0.10 ±0.003 &0.07 ±0.011 &0.14 ±0.001 &0.34 ±0.018 &0.39 ±0.001 &0.02 ±0.001 \\\midrule
    TAG &2 &14.7 ±0.8 &4317 ±229 &1.49 ±0.025 &0.09 ±0.002 &0.23 ±0.008 &0.13 ±0.002 &0.10 ±0.002 &0.07 ±0.005 &0.14 ±0.001 &0.33 ±0.011 &0.39 ±0.001 &0.02 ±0.001 \\
    \textbf{MuFL} &2 &6.0 ±0.1 &1986 ±108 &1.45 ±0.021 &0.08 ±0.003 &0.22 ±0.008 &0.12 ±0.001 &0.10 ±0.001 &0.06 ±0.004 &0.14 ±0.000 &0.32 ±0.011 &0.39 ±0.001 &0.02 ±0.001 \\\midrule
    TAG &3 &16.5 ±2.6 &4854 ±751 &1.44 ±0.014 &0.09 ±0.006 &0.23 ±0.009 &0.12 ±0.001 &0.10 ±0.002 &0.03 ±0.004 &0.14 ±0.000 &0.33 ±0.009 &0.39 ±0.001 &0.02 ±0.000 \\
    \textbf{MuFL} &3 &6.6 ±0.4 &1955 ±104 &1.39 ±0.030 &0.07 ±0.005 &0.22 ±0.008 &0.12 ±0.002 &0.08 ±0.002 &0.05 ±0.003 &0.14 ±0.001 &0.32 ±0.011 &0.38 ±0.001 &0.02 ±0.000 \\\midrule
    TAG &4 &15.8 ±2.4 &4639 ±717 &1.44 ±0.007 &0.07 ±0.003 &0.24 ±0.002 &0.11 ±0.001 &0.10 ±0.002 &0.03 ±0.004 &0.14 ±0.000 &0.35 ±0.003 &0.39 ±0.001 &0.02 ±0.000 \\
    \textbf{MuFL} &4 &7.5 ±0.3 &2201 ±94 &1.40 ±0.027 &0.06 ±0.004 &0.22 ±0.008 &0.12 ±0.003 &0.08 ±0.002 &0.05 ±0.001 &0.14 ±0.001 &0.32 ±0.011 &0.39 ±0.001 &0.02 ±0.001 \\\midrule
    \textbf{MuFL} &5 &8.3 ±0.4 &2439 ±105 &1.40 ±0.028 &0.06 ±0.004 &0.22 ±0.008 &0.12 ±0.003 &0.08 ±0.002 &0.05 ±0.000 &0.14 ±0.002 &0.32 ±0.011 &0.39 ±0.001 &0.02 ±0.001 \\
    \bottomrule
    \end{tabular}
  \end{table}
\endgroup


\section{Additional Experimental Evaluation}
\label{apx:evaluation}

\begin{wrapfigure}{r}{.4\textwidth}
  \vspace{-1cm}
  \centering
  \includegraphics[width=0.4\textwidth]{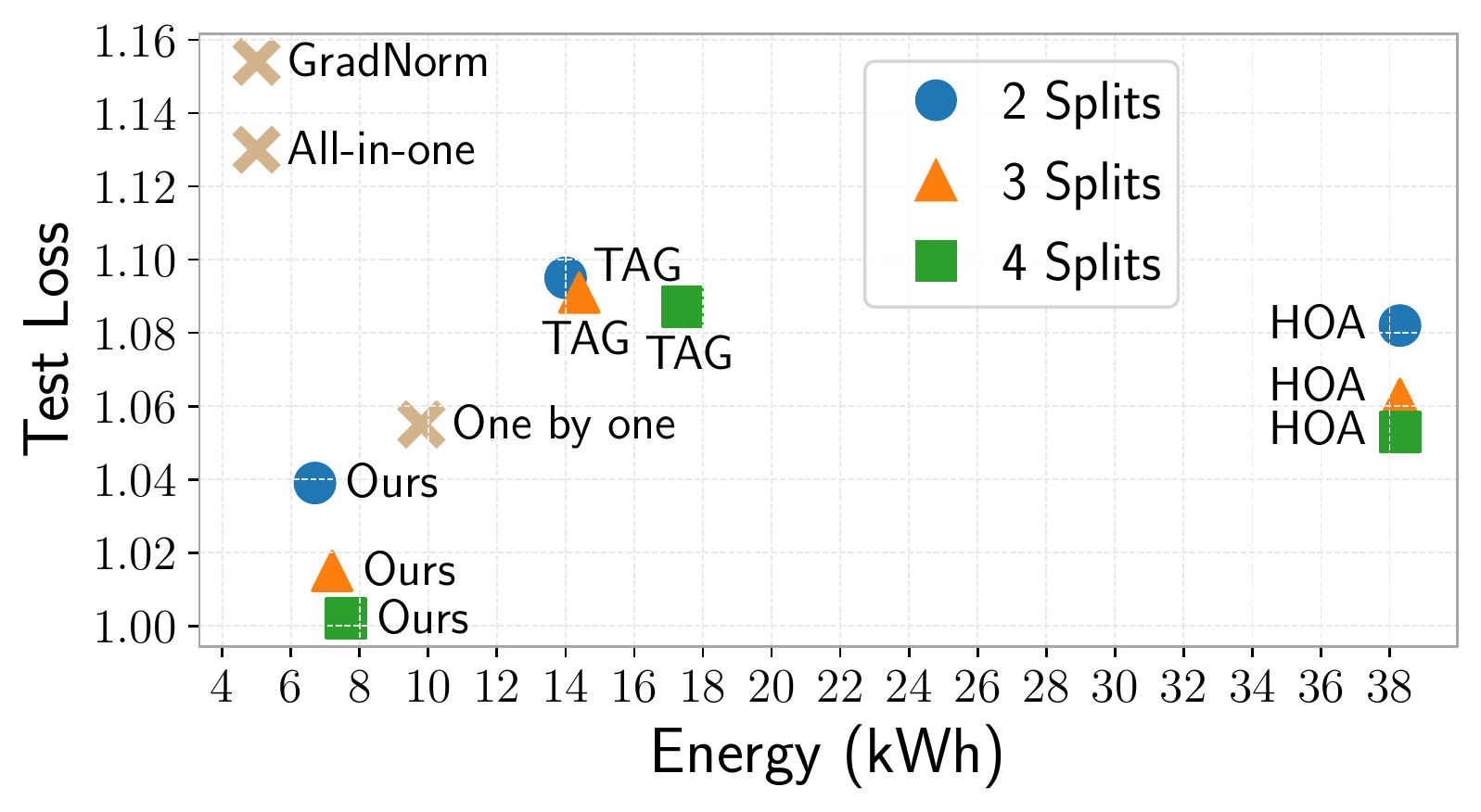}
  \caption{Compare test loss and energy consumption on activity set \texttt{erckt}.}
  \label{fig:erckt-resource}  
  \vspace{-1cm}
\end{wrapfigure}

This section provides more experimental results, including comprehensive results of performance evaluation and additional ablation studies.

\subsection{Performance Evaluation}


Table \ref{tab:sdnkt-full} and \ref{tab:sdnkterca-full} provide comprehensive comparison of different methods on test loss and energy consumption on activity sets \texttt{sdnkt} and \texttt{sdnkterca}, respectively. They complement the results in Figure \ref{fig:performance-vs-resource}. Besides, Table \ref{tab:erckt-full} and Figure \ref{fig:erckt-resource} compares these methods on activity set \texttt{erckt}. The results on \texttt{erckt} is similar to results on the other activity sets; our method achieves the best performance with around 40\% less energy consumption than the one-by-one method and with slightly more energy consumption than all-in-one methods.

\begin{figure}[t]
  \centering
  \begin{subfigure}[t]{0.32\textwidth}
    \includegraphics[width=\textwidth]{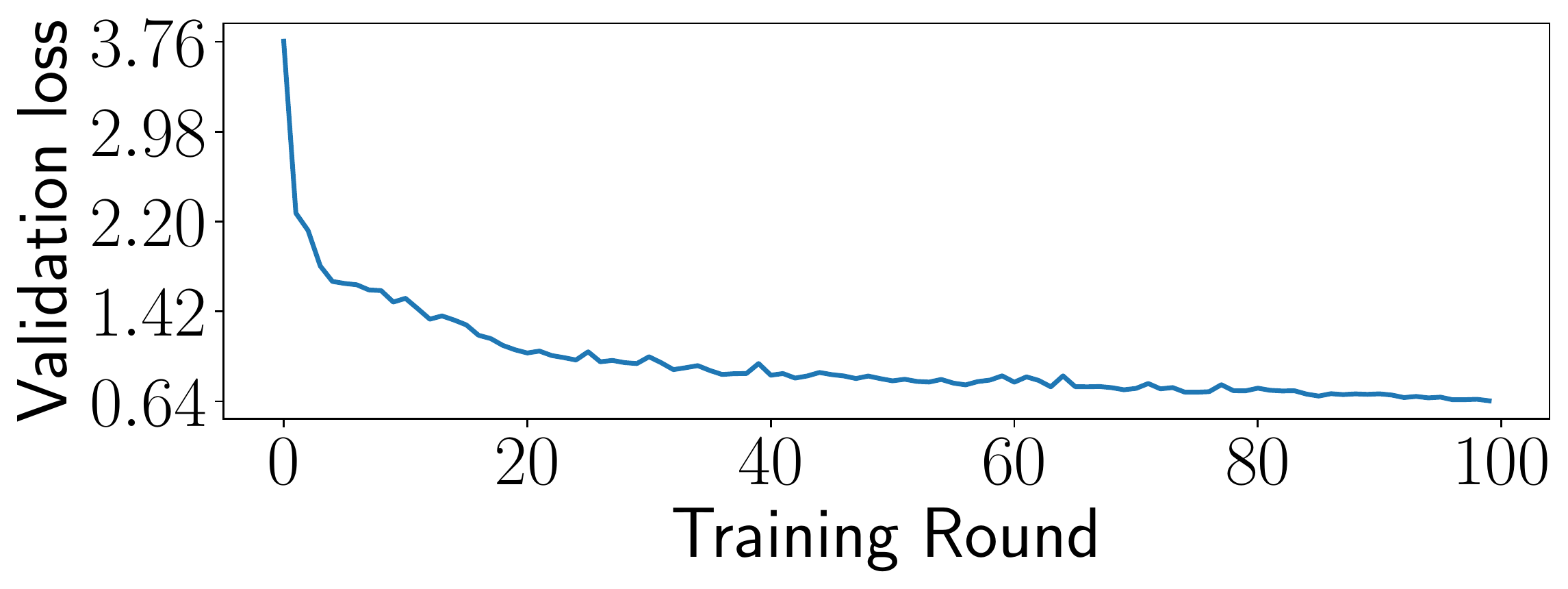}
    \caption{Activity set \texttt{sdnkt}}
    \label{fig:sdnkt-loss}
  \end{subfigure}
  \hfill    
  \begin{subfigure}[t]{0.32\textwidth}
    \includegraphics[width=\textwidth]{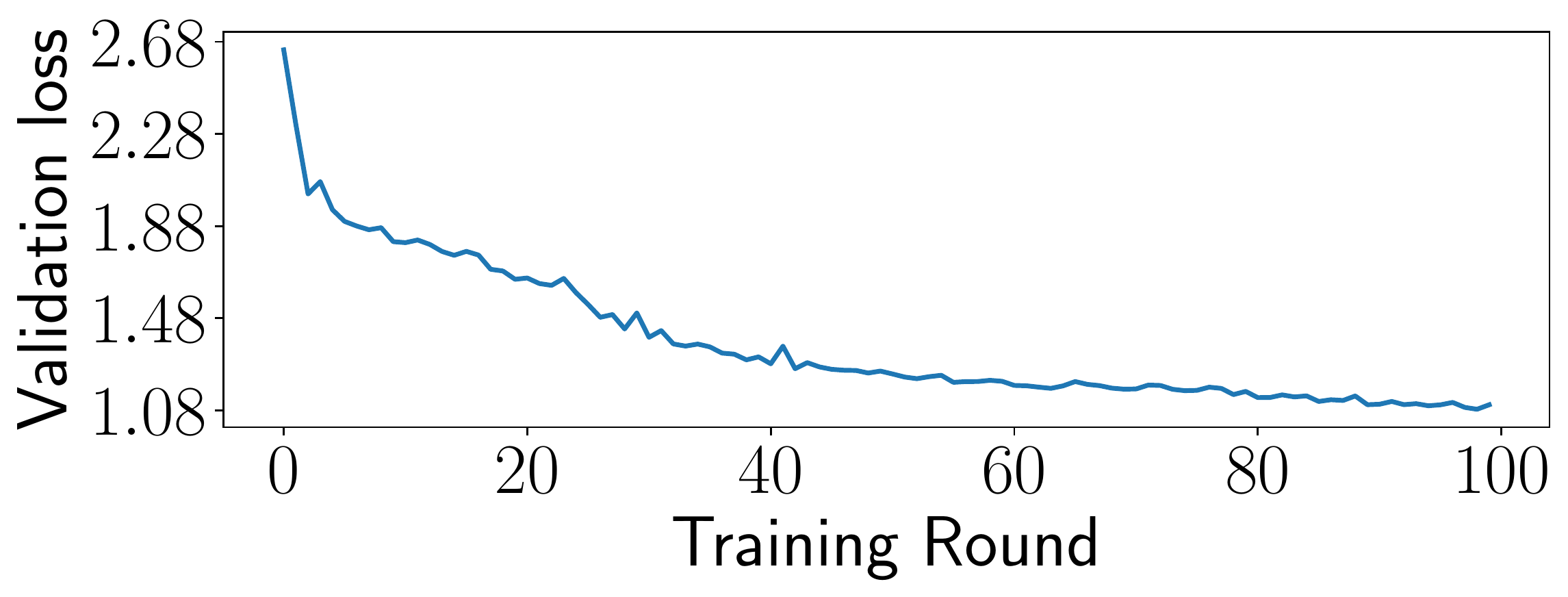}
    \caption{Activity set \texttt{erckt}}
    \label{fig:erckt-loss}
  \end{subfigure}  
  \hfill    
  \begin{subfigure}[t]{0.32\textwidth}
    \includegraphics[width=\textwidth]{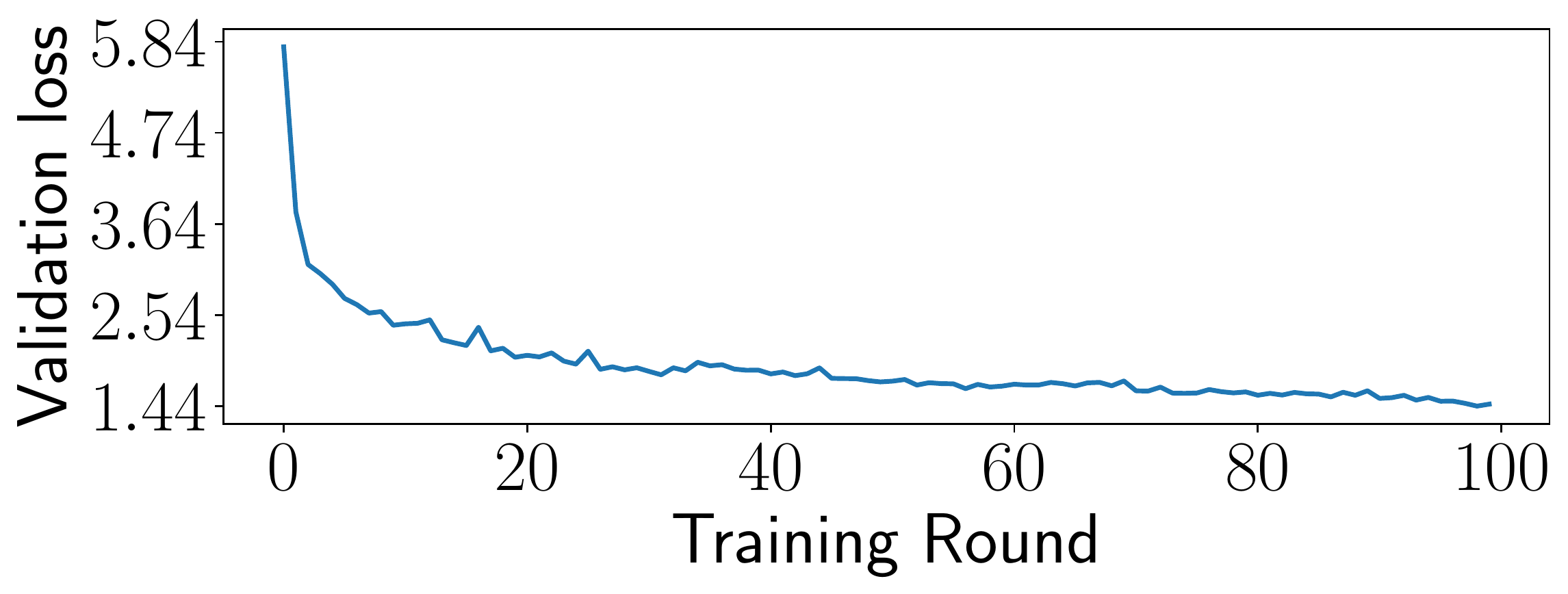}
    \caption{Activity set \texttt{sdnkterca}}
    \label{fig:sdnkterca-loss}
  \end{subfigure}    
 \caption{Changes of validation loss over the course of training on activity sets: (a) \texttt{sdnkt}, (b) \texttt{erckt}, and (c) \texttt{sdnkterca}. Validation loss converges as training proceeds.}
 \label{fig:validation-losses}
\end{figure}

\begin{table}[t!]\centering
  \caption{Activity splitting results of TAG \cite{fifty2021tag} and MuFL on activity sets \texttt{sdnkt}, \texttt{erckt}, and \texttt{sdnkterca}. Activities of each split is separated by a comma.}\label{tab:splitting-results}
  \begin{tabular}{cccccc}\toprule
  Method &Activity Set &Two Splits &Three Splits &Four Splits &Five Splits \\\midrule
  TAG &\texttt{sdnkt} &sdn,kt &sd,dn,kt &sd,sdn,dn,kt &- \\
  MuFL &\texttt{sdnkt} &sdn,kt &sdn,k,t &sd,n,k,t &s,d,n,k,t \\\midrule
  TAG &\texttt{erckt} &er,rckt &er,kt,rc &er,kt,rc,rt &- \\
  MuFL &\texttt{erckt} &er,ckt &er,c,kt &er,c,k,t &e,r,c,k,t \\\midrule
  TAG &\texttt{sdnkterca} &sdnkterca,dr &sdnerc,dr,kta &sc,dr,ne,kta &- \\
  MuFL &\texttt{sdnkterca} &snkteac,dr &snec,dr,kta &sn,dr,ka,etc &sn,dr,ka,e,tc \\
  \bottomrule
  \end{tabular}
\end{table}

Additionally, Table \ref{tab:sdnkt-full}, \ref{tab:erckt-full} and \ref{tab:sdnkterca-full} also provide carbon footprints (CO2eq) of different methods. The carbon footprints are estimated using Carbontracker \cite{anthony2020carbontracker}.\footnote{Carbon intensity of a training varies over geographical regions according to \cite{anthony2020carbontracker}. We use the national level (the United Kingdom as the default setting of the tool) of carbon intensity for a fair comparison across different methods. These carbon footprints serve as a proxy for evaluation of the actual carbon emissions.} Our method reduces around 40\% on carbon footprints on these three activity sets compared with one-by-one training; it reduces 1526gCO$_2$eq or equivalent to traveling 12.68km by car on \texttt{sdnkterca}. The reduction is even more significant when compared with TAG and HOA. Although we run experiments using Tesla V100 GPU, the relative results of energy and carbon footprint among different methods should be representative of the scenarios of edge devices.

\begin{table}[t]\centering
  \caption{Results of the optimal and worst splits in three runs of experiments. They are not identical due to variances in three runs of experiments.}\label{tab:optimal-worst-splits}
  \begin{tabular}{c|c|ccc|ccc}\toprule
  Activity Set &Splits &\multicolumn{3}{c|}{Optimal Splits} &\multicolumn{3}{c}{Worst Splits} \\\midrule
  \multirow{2}{*}{\texttt{sdnkt}} &2 &dk,snt &sn,dkt &nt,sdk &st,dnk &st,dnk &st,dnk \\
  &3 &t,sn,dk &k,t,sdn &d,sn,kt &d,st,nk &d,st,nk &s,dt,nk \\\midrule
  \multirow{2}{*}{\texttt{erckt}} &2 &r,eckt &t,erck &et,rck &rk,ect &ek,rct &e,rckt \\
  &3 &r,ec,kt &r,t,eck &r,ec,kt &c,e,rk &e,k,rct &e,rt,ck \\
  \bottomrule
  \end{tabular}
\end{table}

\begin{figure}[t]
  \centering
  \begin{subfigure}[t]{0.31\textwidth}
    \includegraphics[width=\textwidth]{charts/affinity_onto_s.pdf}
    \caption{Affinities to activity \texttt{s}}
  \end{subfigure}
  \hfill    
  \begin{subfigure}[t]{0.31\textwidth}
    \includegraphics[width=\textwidth]{charts/affinity_onto_d.pdf}
    \caption{Affinities to activity \texttt{d}}
  \end{subfigure}  
  \hfill    
  \begin{subfigure}[t]{0.31\textwidth}
    \includegraphics[width=\textwidth]{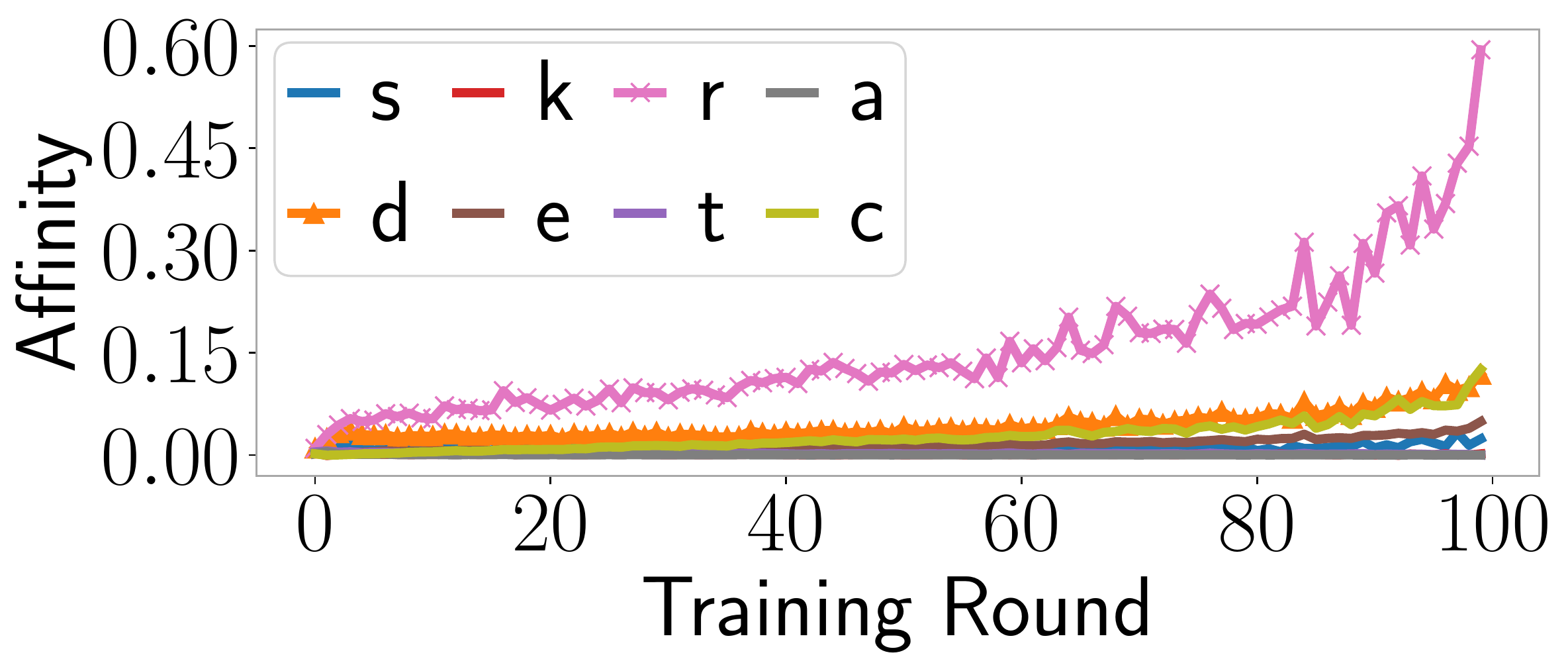}
    \caption{Affinities to activity \texttt{n}}
  \end{subfigure}    
  \hfill
  \begin{subfigure}[t]{0.31\textwidth}
    \includegraphics[width=\textwidth]{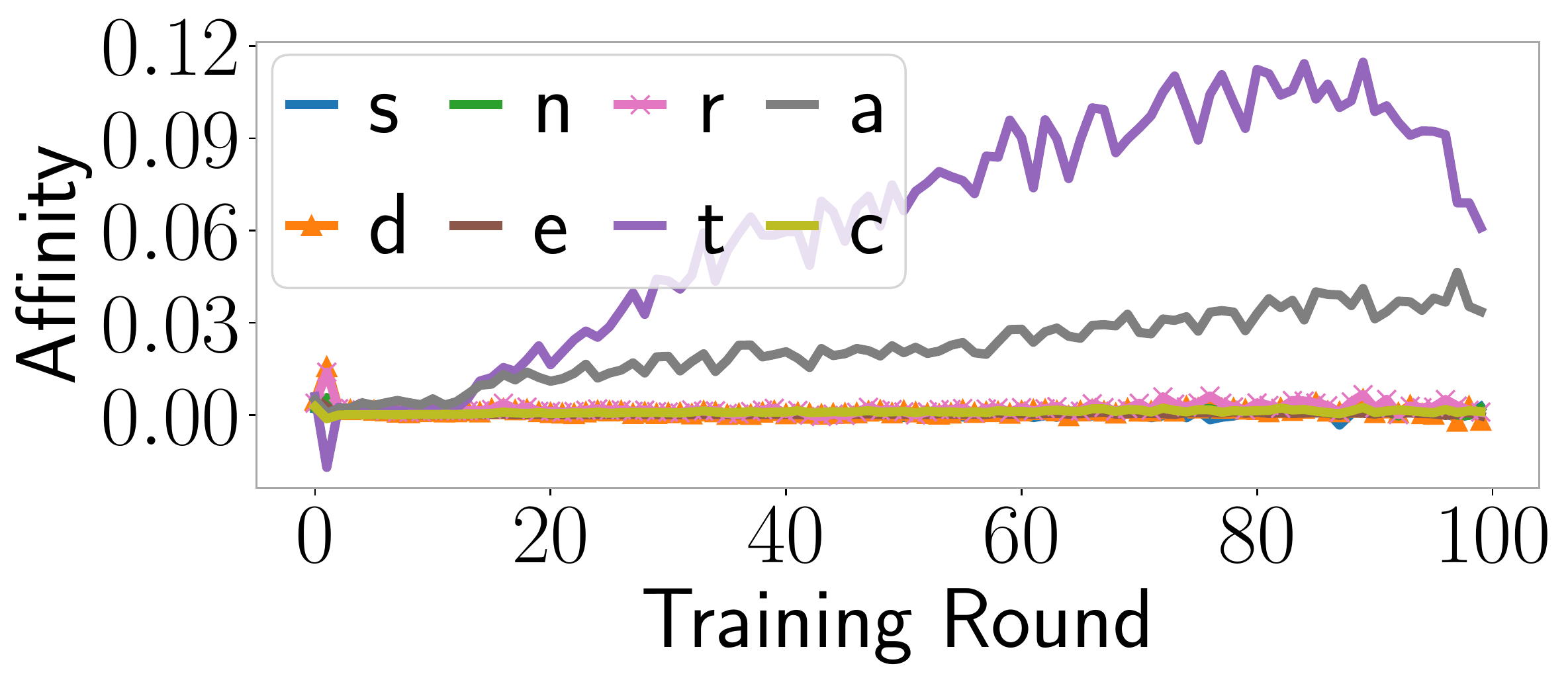}
    \caption{Affinities to activity \texttt{k}}
  \end{subfigure}
  \hfill    
  \begin{subfigure}[t]{0.31\textwidth}
    \includegraphics[width=\textwidth]{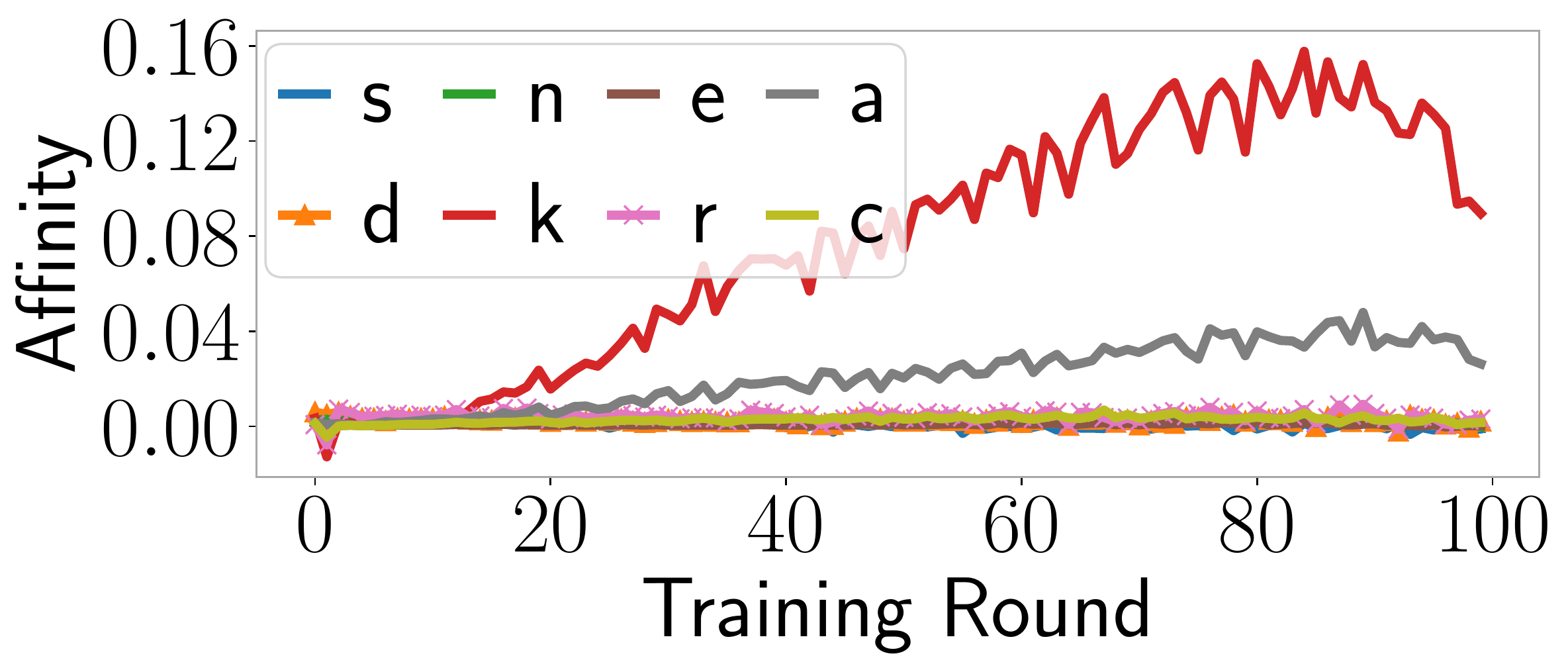}
    \caption{Affinities to activity \texttt{t}}
  \end{subfigure}  
  \hfill    
  \begin{subfigure}[t]{0.31\textwidth}
    \includegraphics[width=\textwidth]{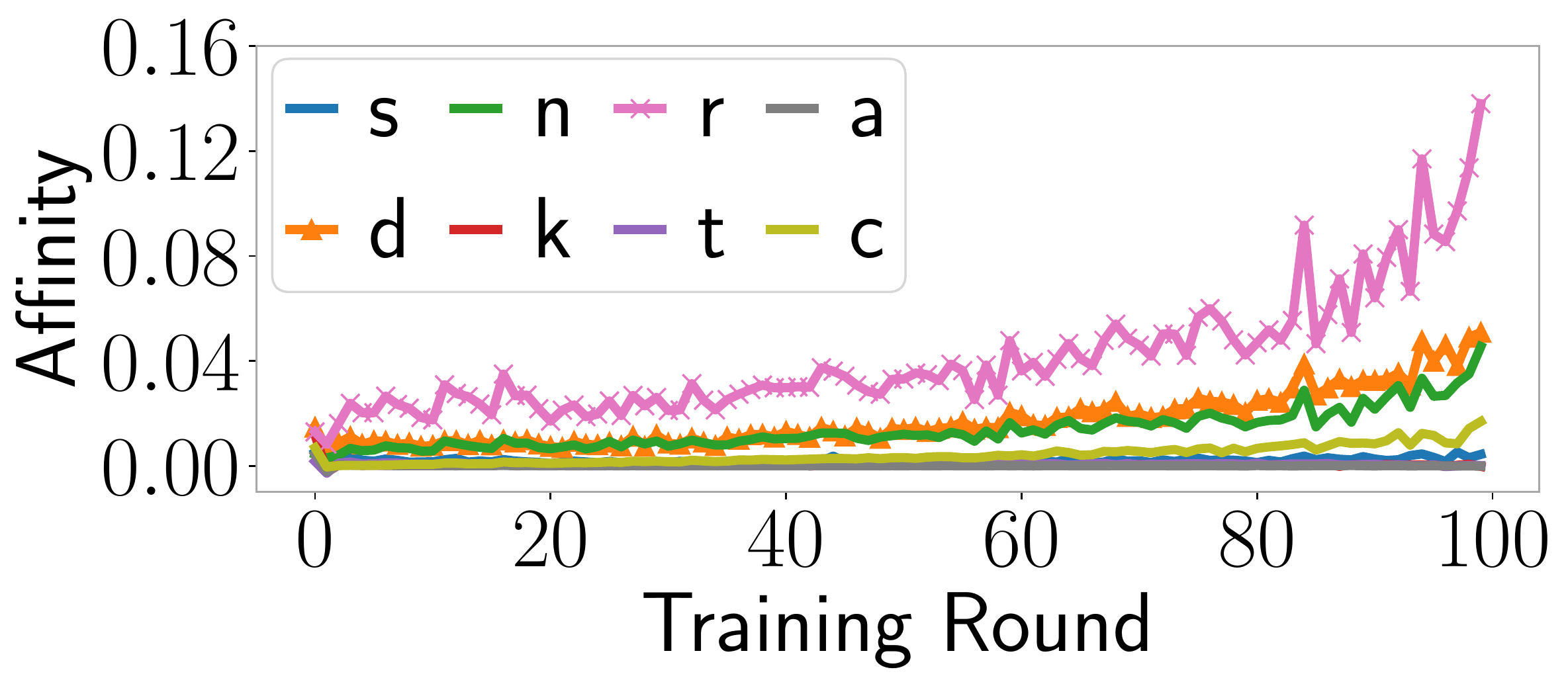}
    \caption{Affinities to activity \texttt{e}}
  \end{subfigure}    
  \hfill
  \begin{subfigure}[t]{0.31\textwidth}
    \includegraphics[width=\textwidth]{charts/affinity_onto_r.pdf}
    \caption{Affinities to activity \texttt{r}}
  \end{subfigure}
  \hfill    
  \begin{subfigure}[t]{0.31\textwidth}
    \includegraphics[width=\textwidth]{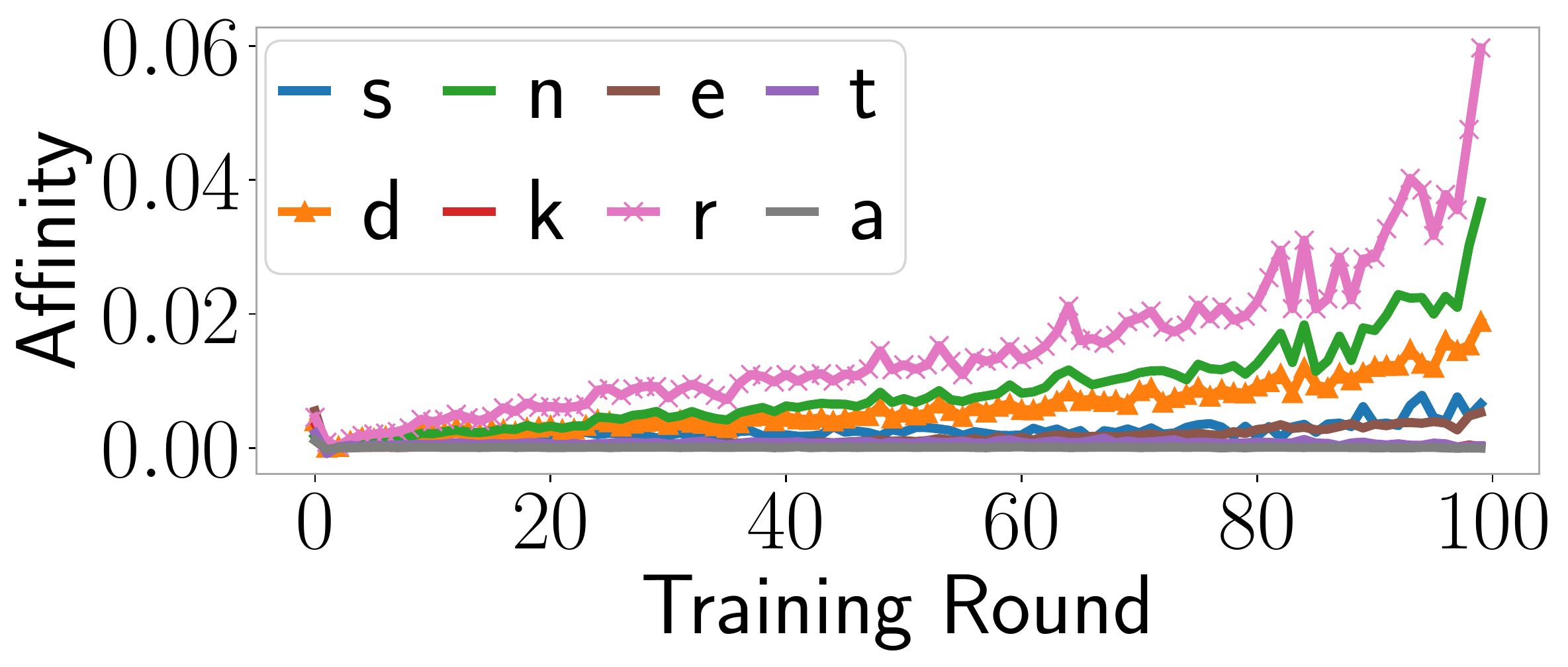}
    \caption{Affinities to activity \texttt{c}}
  \end{subfigure}  
  \hfill    
  \begin{subfigure}[t]{0.31\textwidth}
    \includegraphics[width=\textwidth]{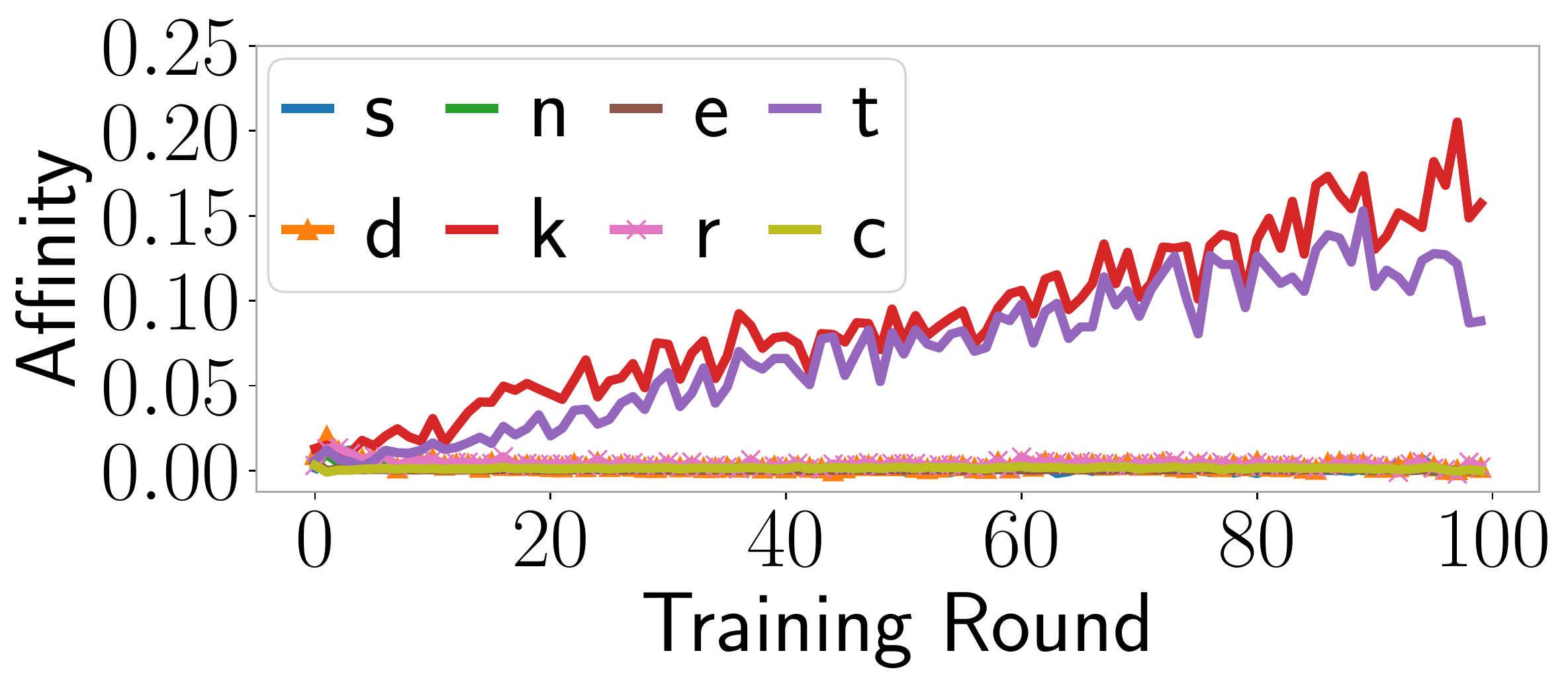}
    \caption{Affinities to activity \texttt{a}}
  \end{subfigure}    
  \hfill
 \caption{Changes of affinity scores of one activity to the other over the course of training on activity set \texttt{sdnkterca}. The trends of affinities emerge at the early stage of training.}
 \label{fig:affinity-full}
\end{figure}

\begin{figure}[t]
  \centering
  \begin{subfigure}[t]{0.29\textwidth}
    \includegraphics[width=\textwidth]{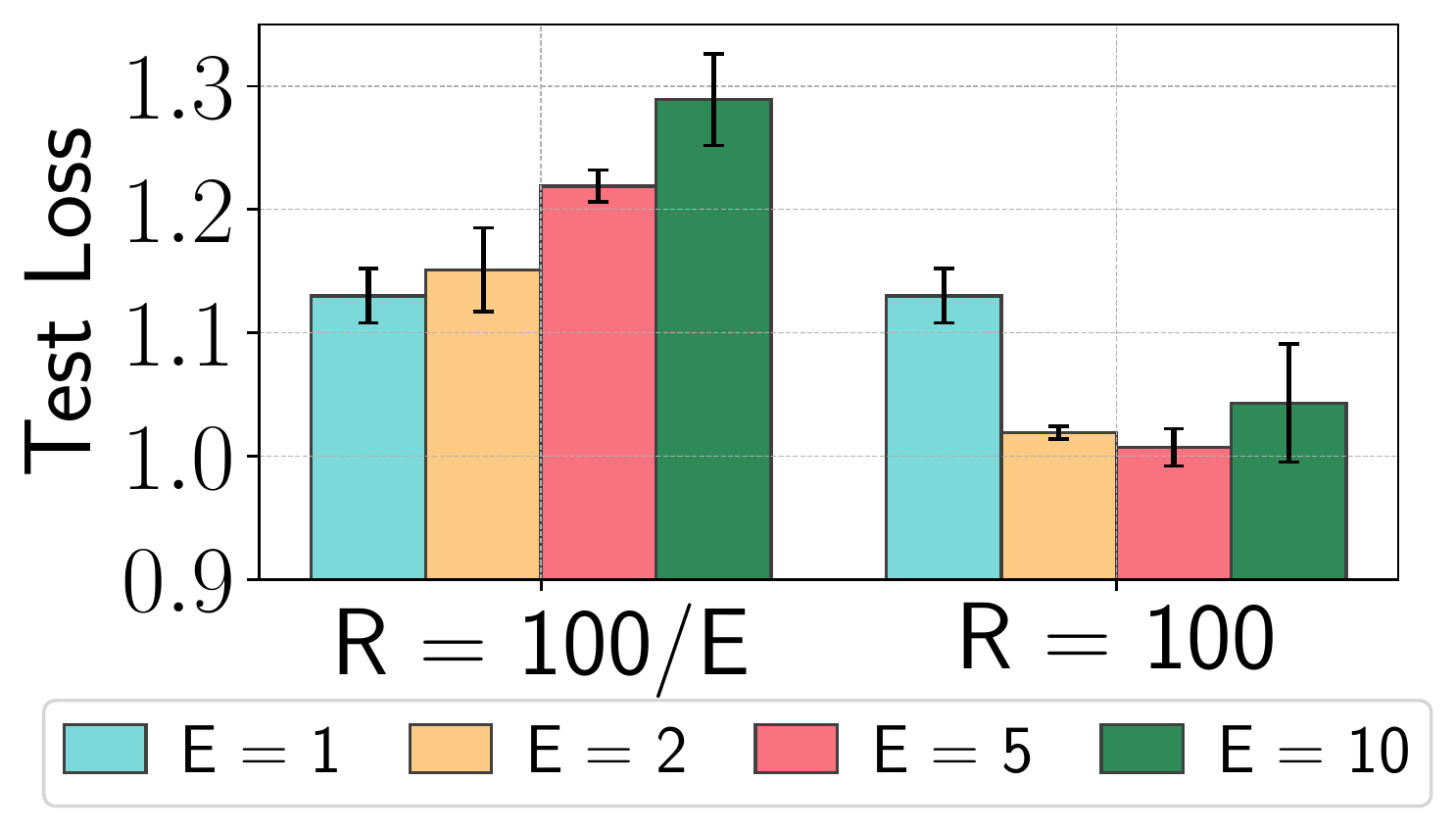}
    \caption{Impact of $E$ on \texttt{erckt}}
    \label{fig:erckt-local-epoch}
\end{subfigure}
\begin{subfigure}[t]{0.29\textwidth}
  \includegraphics[width=\textwidth]{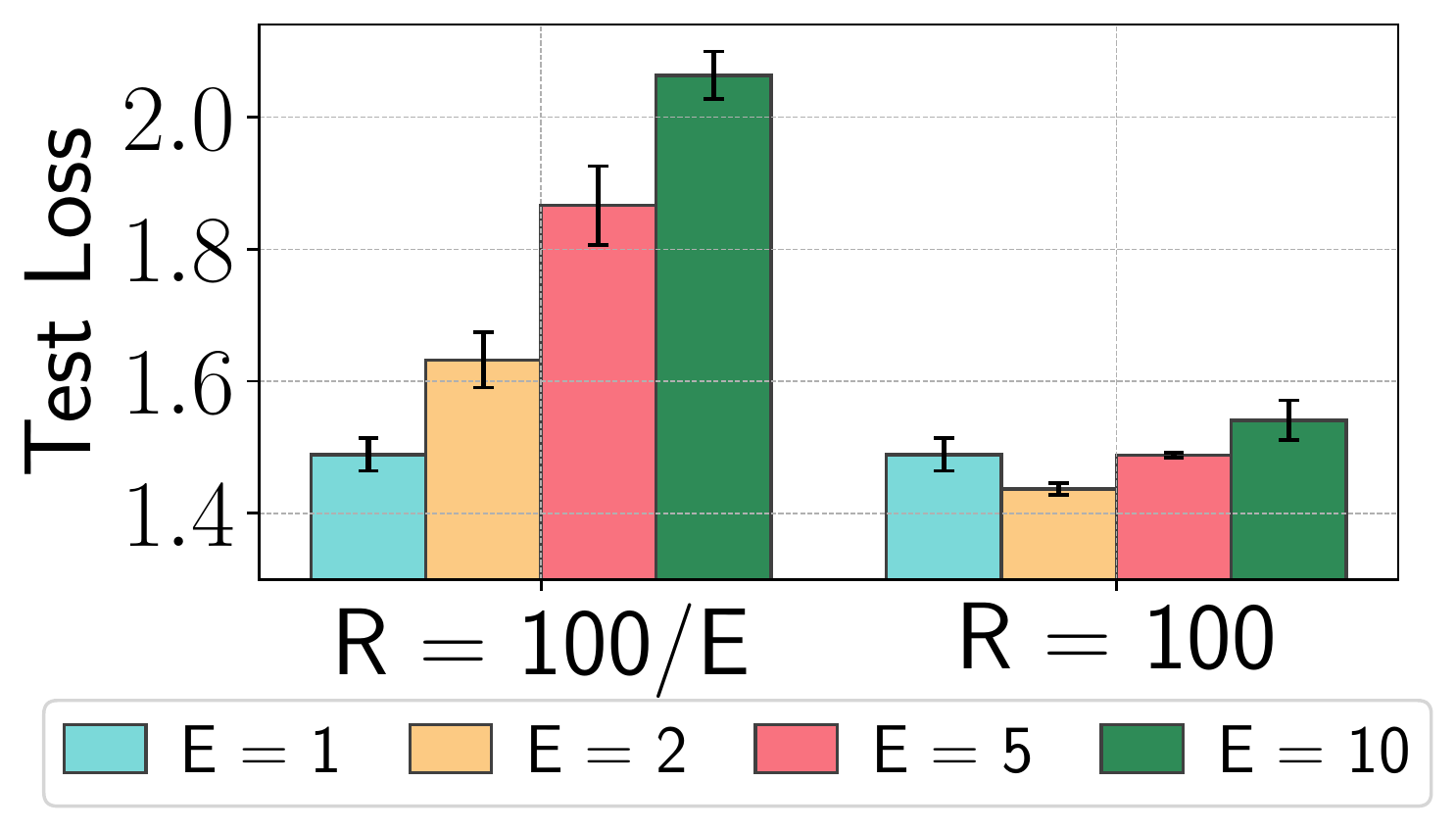}
  \caption{Impact of $E$ on \texttt{sdnkterca}}
  \label{fig:sdnkterca-local-epoch}
\end{subfigure}
\begin{subfigure}[t]{0.4\textwidth}
  \includegraphics[width=\textwidth]{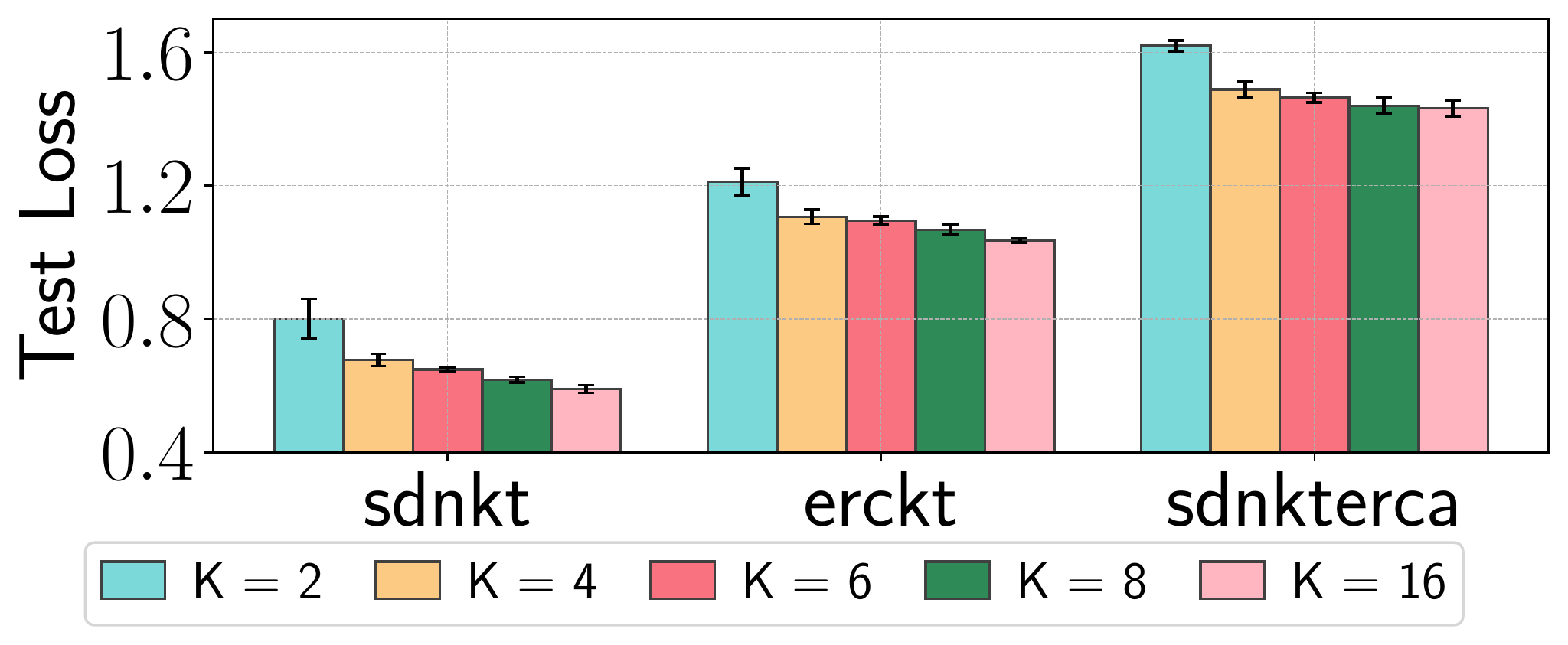}
  \caption{Impact of $K$}
  \label{fig:num-clients-full}
\end{subfigure}
 \caption{Ablation studies on the impact of local epoch $E$ and impact of the number of selected clients $K$. Larger $E$ degrades performance when $R = \nicefrac{100}{E}$, but it could reduce losses when $R = 100$. Larger $E$ (with $R=100$) and $K$ requires higher computation. They could reduce losses, but the marginal benefit decreases as computation increases.}
 \label{fig:abl}

\end{figure}


\subsection{Additional Analysis and Ablation Studies}

This section presents additional analysis of MuFL and provides additional ablation studies.

\paragraph{Changes of Vadiation Loss} Figure \ref{fig:validation-losses} presents validation losses over the course of all-in-one training of three training activity sets \texttt{sdnkt}, \texttt{erckt}, and \texttt{sdnkterca}. It shows that validation losses converge as training proceeds.

\paragraph{Splitting Results of Various Methods} We provide results of activity splitting of TAG \cite{fifty2021tag} and MuFL in Table \ref{tab:splitting-results}. For hierarchical splitting, they further split into three splits from the results of two splits. In particular, the results of hierarchical splitting of \texttt{erckt} and \texttt{sdnkterca} are the same as their three splits. The hierarchical splitting result of \texttt{sdnkt} is from \{\texttt{sdn,kt}\} to \{\texttt{sd,n,kt}\} as the hierarchical splitting further divides the split with more training activities (sdn).  

Besides, Table \ref{tab:optimal-worst-splits} presents the splitting results of the optimal and worst splits. They are not identical due to variances in multiple runs of experiments. We report the mean and standard deviation of test losses of the optimal splits and the worst splits in Table \ref{tab:comparison-optimal-worst}. The large variances of the optimal and worst splits suggest the instability of splitting by measuring the performances of training from scratch in the FL settings and show the advantage of our methods in obtaining stable splits.

\begin{table}[t]\centering
  \caption{Comparison of test loss using different numbers of selected clients $K$. MuFL achieves even better performance on $K = 8$.}\label{tab:8-client-mufl}
  \begin{tabular}{ccccccccc}\toprule
  &K &Total Loss &s &d &n &k &t \\\midrule
  All-in-one &4 &0.677 &0.087 &0.246 &0.136 &0.126 &0.083 \\
  All-in-one &8 &0.618 &0.076 &0.227 &0.130 &0.109 &0.077 \\
  MuFL (two splits) &4 &0.578 &0.069 &0.231 &0.124 &0.102 &0.052 \\
  MuFL (two splits) &8 &\textbf{0.512} &\textbf{0.060} &\textbf{0.202} &\textbf{0.117} &\textbf{0.083} &\textbf{0.048} \\
  \bottomrule
  \end{tabular}
\end{table}

\begin{figure}[t!]
  \centering
    \subfloat[Test loss distribution of standalone training]{\label{fig:loss-distribution}%
        \includegraphics[width=0.42\linewidth]{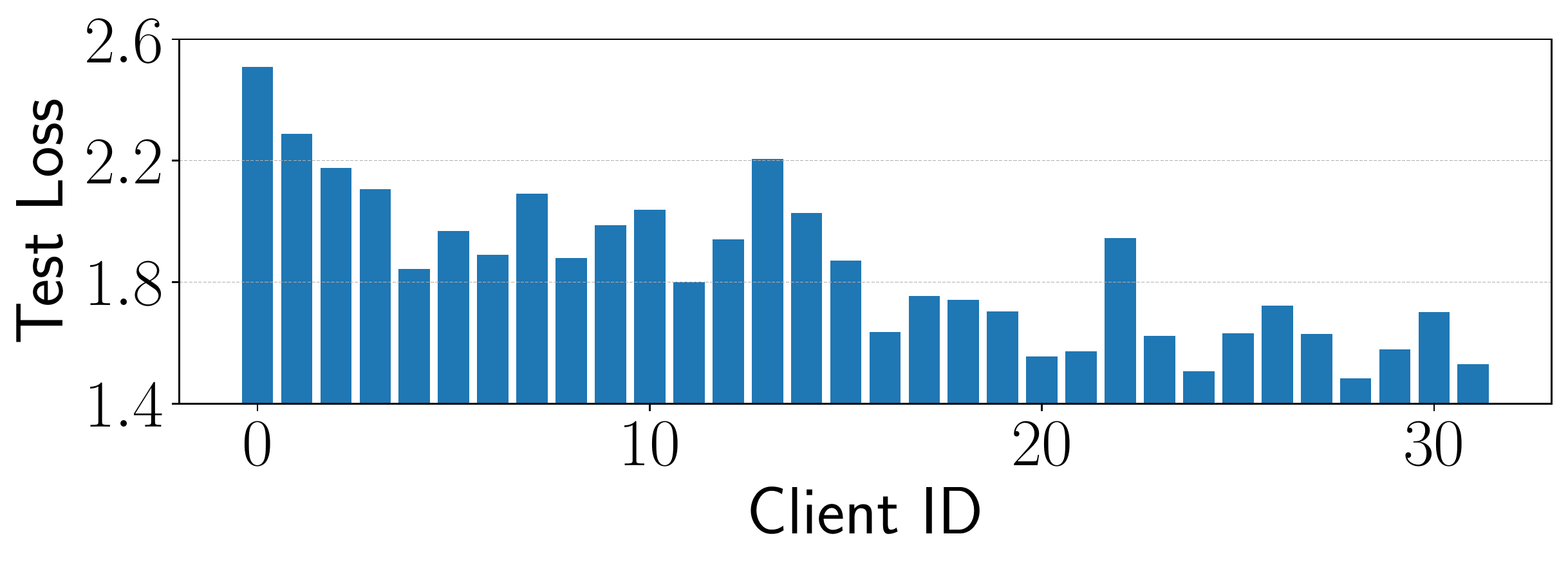}
    }
    \hspace{0.5cm}
    \subfloat[Test loss comparison]{\label{fig:standalone-comparison}%
      \begin{tabular}{cc}\toprule
        Methods &Test Loss \\\midrule
        Standalone &1.842 ± 0.248 \\
        All-in-one &0.677 ± 0.018 \\
        MuFL &0.548 ± 0.001 \\
        \bottomrule
        \end{tabular}
    }
  \caption{Performance (test loss) of standalone training that conducts training using data in each client independently: (a) shows the test loss distribution of these thirty-two clients. (b) compares test losses of standalone training and FL methods. We run the experiments on activity set \texttt{sdnkt}.}
  \label{fig:standalone}
\end{figure}

\paragraph{Dataset Size and Performance} The dataset size of activity set \texttt{sdnkt} is around 315GB in our experiments, compared to 2.4TB of dataset used in experiments of TAG \cite{fifty2021tag}. The test loss of ours (0.512 in Table \ref{tab:8-client-mufl}), however, is better than the optimal one in TAG \cite{fifty2021tag} (0.5246). This back-of-the-envelope comparison indicates the potential to extend our approaches to multi-task learning. Besides, it could also suggest that our data size is sufficient for evaluation. 

\paragraph{Impact of Affinity Computation Frequency $f$} The frequency of computing affinities in Equation \ref{eq:single-affinity} determines the amount of extra needed computation. We use $f = 5$ and compute affinities for the first ten rounds for all experiments because the trend of affinities emerge in the early stage of training in Figure \ref{fig:affinity-full}. It would increase the computation of all-in-one training by around 2\%, which is already factored into the energy consumption computation in previous experiments. The results in Table \ref{tab:sdnkt-full}, \ref{tab:erckt-full}, and \ref{tab:sdnkterca-full} show that MuFL is effective with this setting and the amount of computation is acceptable.

\paragraph{Impact of Local Epoch} Figure \ref{fig:erckt-local-epoch} and \ref{fig:sdnkterca-local-epoch} show the impact of local epoch $E$ on activity sets \texttt{erckt} and \texttt{sdnkterca}, respectively. They complement results of activity set \texttt{sdnkt} in Figure \ref{fig:sdnkt-local-epoch}. On the one hand, larger $E$ degrades performance when $R = \nicefrac{100}{E}$, even though the total computation remains. It could be due to the impact of statistical heterogeneity; larger $E$ amplifies the heterogeneity among selected clients. On the other hand, larger $E$ could lead to better performance when fixing $R = 100$. It is especially effective when increasing $E = 1$ to $E = 2$, but further increasing $E$ could degrade the performance. It indicates that simply increasing computation has limited capability to improve the performance. 

\paragraph{Impact of The Number of Selected Clients} Figure \ref{fig:num-clients-full} compares the performance of different numbers of selected clients $K = \{2, 4, 6, 8, 16\}$ on three activity sets \texttt{sdnkt}, \texttt{erckt}, and \texttt{sdnkterca}. It complements results in Figure \ref{fig:num-clients}. The results on three activity sets are similar; increasing $K$ reduces losses, but the marginal benefit decreases as $K$ increases. 

The majority of experiments in this study are conducted with $K = 4$. We next analyze the impact of $K$ in MuFL with results of two splits on activity set \texttt{sdnkt} in Table \ref{tab:8-client-mufl}. The results indicate that MuFL is still effective with $K = 8$, which outperforms $K = 4$ and all-in-one training. 

\paragraph{Standalone Training} Standalone training refers to training using data of each client independently. Figure \ref{fig:loss-distribution} shows the test loss distribution of thirty-two clients used in experiments. The client ID corresponds to the dataset size distribution in Figure \ref{fig:client-stats}. These results suggest that larger data sizes of clients may not lead to higher performance. Figure \ref{fig:standalone-comparison} compares test losses of standalone training and federated learning methods. Either all-in-one or our MuFL greatly outperforms standalone training. It suggests the significance of federated learning when data are not sharable among clients. 

\newpage

\section{Algorithm}
\label{apx:algorithm}

\begin{algorithm}[H]
  \centering
  \caption{Our Proposed Smart Multi-tenant FL System (MuFL)}\label{algo:mufl}
  \begin{algorithmic}[1]
    \State \textbf{Input:} training activities $
    \mathcal{A} = \{\alpha_1,\alpha_2,\dots,\alpha_n\}$, a set of available clients $\mathcal{C}$, number of selected clients $K$, local epoch $E$, aggregation weight of client $k$ $p_k$, total training rounds $R$, all-in-one training rounds $R_0$, frequency of computing affinities $f$, batch size $B$
    \State \textbf{Output:} models $\mathcal{W} = \{\omega_1,\omega_2,\dots,\omega_n\}$
    \State 
    \State \textbf{\underline{ServerExecution:}}
    \State Consolidate $\mathcal{A}$ into $\alpha_0$ with a multi-task model $\nu^0 = \{\theta_s\} \cup \{\theta_{\alpha_i}\vert \alpha_i \in \mathcal{A} \}$
    \State Initialize $\nu^0$, i.e. initialize $\omega_i = \{\theta_s\} \cup \{\theta_{\alpha_i}\}$ for $i \in \mathcal{N}=\{1, 2, \dots, n\}$
    \For{\textit{each round} $r = 0, 1, ..., R_0-1$}
       \State $\mathcal{C}^r \gets$ (Randomly select K clients from $\mathcal{C}$)
       \For{\textit{client} $k \in \mathcal{C}^r$ \textit{in parallel}}
           \State $\nu^{r,k}, \hat{\mathcal{S}}_{\alpha_i \rightarrow \alpha_j}^{r,k} \gets$Client($\nu^r$, $\mathcal{A}$, $f$)
       \EndFor
       \State $\nu^{r+1} \gets \sum\limits_{k \in \mathcal{C}^r} p_k \nu^{r,k}$
       \State $\hat{\mathcal{S}}^r_{\alpha_i \rightarrow \alpha_j} \gets \frac{1}{K} \sum\limits_{k \in \mathcal{C}^r} \hat{\mathcal{S}}^{r,k}_{\alpha_i \rightarrow \alpha_j}$
    \EndFor

    \State Compute the values of $\mathcal{S}^r_{\alpha_i \rightarrow \alpha_i}$, $\forall \alpha_i \in \mathcal{A}$, using Equation \ref{eq:self-affinity} 
    \State Compute a set $I \subseteq \{1, 2, 3, ..., m\}$ for splits $\{\mathcal{A}_j | j \in I \}$ using branch-and-bound algorithm and affinites $\mathcal{S}^r_{\alpha_i \rightarrow \alpha_j}$ ($\forall \alpha_i \in \mathcal{A}$ and $\forall \alpha_j \in \mathcal{A}$) to maximize $\mathcal{S}^r_{\alpha_i}$  
    
    \For{\textit{each element $j \in I$}}
      \State Initialize $\nu_j = \{\theta_s^j\} \cup \{\theta_{\alpha_i} | \alpha_i \in \mathcal{A}_j \}$ with parameters of $\nu^{R_0}$ 
      \For{\textit{each round} $r = 0, 1, ..., R - R_0-1$}
        \State $\mathcal{C}^r \gets$ (Random select K from $\mathcal{C}$)
        \For{\textit{client} $k \in \mathcal{C}^r$ \textit{in parallel}}
            \State $\nu_j^{r,k},$ \_ $\gets$Client($\nu_j^r$, $\mathcal{A}_j$, 0)
        \EndFor
        \State $\nu_j^{r+1} \gets \sum\limits_{k \in \mathcal{C}^r} p_k \nu_j^{r,k}$
      \EndFor
    \EndFor
    
    \State Reconstruct $\mathcal{W} = \{\omega_1, \omega_2, \dots, \omega_n\}$ from $\{\nu_j|j \in I\}$ by matching training activities 

    \State \textbf{Return}  $\mathcal{W}$
    \State 

   \State\underline{\textbf{Client}} ($\nu$, $\mathcal{A}$, $f$):
   \State $T = \lfloor\frac{B}{f}\rfloor$ \textbf{if} $f \neq 0$ \textbf{else} $0$
   \For{\textit{local epoch} $e = 1, ..., E$}
      \State Update model parameters $\nu$ with respect to training activities $\mathcal{A}$
       \For{each time-step $t = 1, 2, ..., T$ (every $f$ batches)}
          \State $\forall \alpha_i \in \mathcal{A}$ and $\forall \alpha_j \in \mathcal{A}$, compute affinities of $\mathcal{S}^t_{\alpha_i \rightarrow \alpha_j}$ using Equation \ref{eq:single-affinity} 
       \EndFor
   \EndFor
   \State $\hat{\mathcal{S}}_{\alpha_i \rightarrow \alpha_j} = \frac{1}{ET} \sum\limits_{e=1}^{E} \sum\limits_{t=1}^{T} \mathcal{S}^{t}_{\alpha_i \rightarrow \alpha_j}$, $\forall \alpha_i \in \mathcal{A}$ and $\forall \alpha_j \in \mathcal{A}$

   \State \textbf{Return} $\nu$, $\hat{\mathcal{S}}_{\alpha_i \rightarrow \alpha_j}$
  \end{algorithmic}
\end{algorithm}

\end{document}